\documentclass[letterpaper, conference]{ieeeconf}
\IEEEoverridecommandlockouts     
\usepackage{mathrsfs}
\usepackage{amsmath}
\usepackage{amssymb}
\usepackage{amsfonts}

\usepackage{tikz}
\usepackage{pgfplots}
\usetikzlibrary{shapes,arrows,positioning,fit,hobby,backgrounds,calc}
\usepackage{color}
\usepackage{mydef} 

\usepackage{graphicx} 
\usepackage[ruled,vlined,linesnumbered]{algorithm2e}
\usepackage[font=footnotesize]{caption}
\usepackage[font=footnotesize]{subcaption}

\newcommand{\oprocendsymbol}{\hbox{$\bullet$}} 
\newcommand{\oprocend}{\relax\ifmmode\else\unskip\hfill\fi\oprocendsymbol}

\definecolor{colseta}{HTML}{e68193}
\definecolor{colsetb}{HTML}{cb8748}
\definecolor{colsetc}{HTML}{a59042}
\definecolor{colsetd}{HTML}{839840}
\definecolor{colsete}{HTML}{42a25f}
\definecolor{colsetf}{HTML}{449f8c}
\definecolor{colsetg}{HTML}{469ea3}
\definecolor{colseth}{HTML}{4d9fc3}
\definecolor{colseti}{HTML}{9b9fe8}
\definecolor{colsetj}{HTML}{d181e4}
\definecolor{colsetk}{HTML}{e479c0}

\usepackage{array}

\usepackage{cuted}

\definecolor{myblue}{HTML}{3333b3}

 \renewcommand{\baselinestretch}{0.992}
\tikzstyle{block} = [rectangle, draw, fill=blue!20, 
text width=5em, text centered, rounded corners, minimum height=2.5em]
\tikzstyle{bigblock} = [rectangle, draw, fill=blue!20, 
text width=7.5em, text centered, rounded corners, minimum height=2.5em] 
\tikzstyle{longblock} = [rectangle, draw, fill=blue!20, 
text width=8.5em, text centered, rounded corners, minimum height=2.5em]  
\tikzstyle{line} = [draw, -latex']
\tikzstyle{cloud} = [draw, ellipse,fill=red!20, node distance=3cm,
minimum height=2em]

\tikzset{
	robot/.pic = {
		\coordinate (-center) at (0,-1);
		\coordinate (-left) at (-2,-1);
		\coordinate (-right) at (2,-1);
		\coordinate (-top) at (0,1);
		\coordinate (-down) at (0,-3);
		\draw[fill=gray!70, line width=2pt] (0,0) circle (1);
		\draw[fill=gray!30, line width=2pt] (-2,-3) arc[start angle=180, end angle=0, radius=2cm] -- (2,-3) -- (-2,-3) -- cycle;
	}
}		  

\graphicspath{{./},{./figs/},{./figs/bulk/}}

\title{Behaviorally Heterogeneous Multi-Agent Exploration \\ Using Distributed Task Allocation}  
  
  \author{Nirabhra Mandal$^1$,
  Aamodh Suresh$^2$, Carlos Nieto-Granda$^2$, Sonia Mart\'{i}nez$^1$
  \thanks{This work is supported by the ARL grant: W911NF-23-2-0009.}
  \thanks{$^1$ The authors are with the Contextual Robotics Institute
    and the Department of Mechanical and Aerospace Engineering, UC San
    Diego, California, USA. Emails: \{nmandal, soniamd\}@ucsd.edu.}
  \thanks{$^2$The authors are with the U.S. Army Combat Capabilities
    Development Command, Army Research Laboratory, Adelphi, MD,
    USA. Emails: aamodh@gmail.com, carlos.p.nieto2.civ@army.mil.}}

\begin{document}

\maketitle
\thispagestyle{empty}
\pagestyle{empty}

\begin{abstract}
  We study a problem of multi-agent exploration with
  behaviorally heterogeneous robots.  Each robot maps its surroundings
  using SLAM and identifies a set of areas of interest (AoIs) or
  frontiers that are the most informative to explore next.  The robots
  assess the utility of going to a frontier using \textit{Behavioral
    Entropy} (BE) and then determine which frontier to go to via a
  distributed task assignment scheme. We convert the task assignment
  problem into a non-cooperative game and use a distributed algorithm
  (d-PBRAG) to converge to the Nash equilibrium (which we show is the
  optimal task allocation solution). For unknown utility cases, we
  provide robust bounds using approximate rewards. We test our
  algorithm (which has less communication cost and fast convergence) in simulation, where we explore the effect of sensing
  radii, sensing accuracy, and heterogeneity among robotic teams with
  respect to the time taken to complete exploration and path traveled. We observe that having a team of
  agents with heterogeneous behaviors is beneficial.
\end{abstract}


\section{Introduction}

\begin{figure}[tp]
\begin{center}
	\scalebox{0.7}{\begin{tikzpicture}[node distance = 1.6cm, auto]

\node [block] (slam) {SLAM};
\node [longblock, above right=-0.25cm and 1.1cm of slam,align=center] (frontier) {Frontier Extraction and Clustering};
\node [block, below right=-0.25cm and 0.5cm of slam,align=center] (perceived) {Perceived Occupancy};
\node [block, right=0.5cm of perceived] (behentropy) {Behavioral Entropy};
\node [block, right=5.5cm of slam,align=center] (behutility) {Behavioral Utility};
\node [longblock, below left=1.3cm and -2.05cm of behutility] (comman) {Task Allocator};
\node [block, left=0.5cm of comman] (planner) {Navigation Planner};
\node [block, left=0.5cm of planner] (pathfinder) {Path Finder};
\node [block, left=0.5cm of pathfinder] (controller) {Robot Controller};

\pic [scale=0.2, name=robota] at ([xshift=-0.7cm, yshift=1.1cm]slam.west) {robot};
\node [right=0.4cm of robota-center] (namea) {Robot 1}; 

\begin{pgfonlayer}{background}
\node[draw, thick, rounded corners, fill=blue!50!green, fill opacity=0.3, fit=(slam) (frontier) (behutility) (behentropy) (controller) (comman), inner sep=8pt] (boxb) {};
\end{pgfonlayer}

\node [block, below left=3.5cm and -1.3cm of slam] (merger) {Map Merger};

\pic [scale=0.2, name=robotb] at ([xshift=2cm, yshift=0.2cm]merger.east) {robot};
\node [right=0.4cm of robotb-center] (nameb) {Robot 2}; 
\begin{pgfonlayer}{background}
\node[draw, thick, rounded corners, fill=blue!50!green, fill opacity=0.3, fit=(robotb-left) (nameb), inner sep=8pt] (boxb) {};
\end{pgfonlayer}
\pic [scale=0.2, name=robotc] at ([xshift=2cm, yshift=0.2cm]boxb.east) {robot};
\node [right=0.4cm of robotc-center] (namec) {Robot 3}; 
\begin{pgfonlayer}{background}
\node[draw, thick, rounded corners, fill=blue!50!green, fill opacity=0.3, fit=(robotc-left) (namec), inner sep=8pt] (boxc) {};
\end{pgfonlayer}

\draw[->] (slam.east) -- ([xshift=0.2cm]slam.east)node[above right]{$\mathcal{M}$} |- (frontier.west);
\draw[->] (slam.east) -- ([xshift=0.2cm]slam.east) |- (perceived.west);
\draw[->] (perceived.east) -- (behentropy.west);
\draw[->] (frontier.east) -| ([xshift=-2.5mm]behutility.west) -- node[above left=0.75cm and 0.5cm]{$\mathcal{F}$} (behutility.west);
\draw[->] (behentropy.east) -| ([xshift=-2.5mm]behutility.west) -- node[below=0.7cm]{$H_\alpha$} (behutility.west);
\draw[->] (behutility.south) -- node[midway,right]{$\expect{R}$} ([xshift=0.6cm]comman.north);
\draw[<->] (comman.south) |- node[midway,above left]{$\{M^q,S^q\}_{q \in \mathcal{F}}$}([xshift=-2cm, yshift=-0.9cm]comman.south) |- (boxc.west);
\draw[<->] (comman.south) |- ([xshift=-2cm, yshift=-0.9cm]comman.south) |- (boxb.east);
\draw[->] (comman.west) -- node[midway,above]{$\mathcal{T}$}(planner.east);
\draw[->] (planner.west) -- (pathfinder.east);
\draw[<->] (pathfinder.west) -- node[midway,above]{$u$}(controller.east);
\draw[dashed, ->] ([xshift=0.3cm]slam.south) -- node[midway,right]{$x$}([xshift=0.3cm, yshift=-0.5cm]slam.south);
\draw[dashed, ->] ([yshift=0.6cm]behutility.north) -- node[midway,left]{$x$}(behutility.north);
\draw[dashed, ->] ([yshift=0.5cm]planner.north) -- node[midway,left]{$x$}(planner.north);
\draw[<->] ([xshift=-0.3cm]slam.south) -- node[midway,left]{$\map_l$}([xshift=-0.3cm, yshift=-0.8cm]slam.south) -| ([xshift=-1cm]merger.west) -- (merger.west);
\draw[<->] (merger.north) -- ([yshift=0.3cm]merger.north) -| (boxb.north);
\draw[<->] (merger.south) -- ([yshift=-0.3cm]merger.south) -| (boxc.south);
\end{tikzpicture}}
	\caption{Proposed exploration framework with 3 robots in close
          proximity. Proximity determines the communication
          architecture. Each robot shares the same internal logic
          framework as in the teal box. The SLAM system provides a
          local map $\mathcal{M}_l$, a global map $\mathcal{M}$ (using
          an external map merger scheme) and robot localization
          $x$. Then, frontiers $\mathcal{F}$ (clustered using some
          scheme) and perceived occupancy are extracted from
          $\mathcal{M}$ and further Behavioral Entropy $H_\alpha$ is
          calculated for each cluster. This, along with the distance to
          the cluster, gives the sample Behavioral Utility
          $\expect{R}$. Then, utilities are communicated with other
          robots to allocate the appropriate frontiers. The navigation
          planner takes these allocations and finds a traveling
          salesman tour among them. The path finder then finds a path
          to the frontier using RRT, which, in turn, sends global and
          local plans $u$ to the controller.}
	\label{fig:framework}
\end{center}
\end{figure}
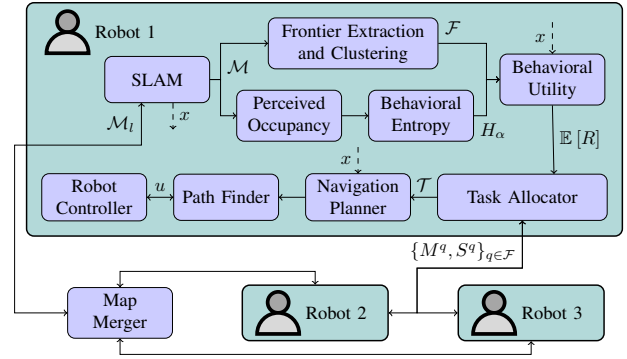

Exploring uncertain, hazardous environments is often handled by
coordinating autonomous robots to improve efficiency, effectiveness,
and robustness towards individual failures. Robot teams exhibiting
diverse behaviors (in terms of assessing risks and rewards) are hence
desired to cater to a wide variety of complex environments and
scenarios.  Typically, exploration frameworks can be broken down into
(see Figure~\ref{fig:framework}): a) perception (robots gather position and
environment knowledge using
SLAM~\cite{HDW-TB:06, TB-HDW:06}), b) reasoning (robots evaluate AoIs---areas
of interest---using available knowledge), c) communication and
assignment (robots discuss and assign tasks or AoIs among themselves),
and d) navigation (robots plan and travel to selected AoIs). Robotic
perception~\cite{YT-YC-FHA-CN-JPH-LC:22, CN-JGR-HIC:14}, and
navigation~\cite{GW:17, SK-AA-NA:21} have been widely studied across
multiple communities.
With regards to decision making, it is evident that, diverse human
teams comprising different behaviors~\cite{DVK-LHN-DJGD:20} fare
better
in situations encouraging synergy.  However, human-inspired reasoning
among robots and its combination with efficient and distributed
assignment schemes remain a nascent, unexplored area of research. Our
framework (outlined in Figure~\ref{fig:framework}) aims to address
this.

Behavioral Entropy~\cite{AS-CN-SM:24-ral} incorporates human-inspired
evaluation of information quality and promotes diversity in
perception, and evaluation. Further, such human nature towards
assessing risk naturally leans into game theoretic territory to deal
with interactions and team decision making. However, since robot
teams often encounter communication limitations (due to distance and
bandwidth constraints); efficient, distributed algorithms are
desired to ensure smooth operation.  In this paper, we investigate if
diversity and heterogeneity coupled with distributed task allocation
subroutines are beneficial robot teams tasked with exploring
an unknown map.

\paragraph*{Literature Review}
There is a rich literature dealing with SLAM techniques for
unknown environment exploration using visual
odometry~\cite{YT-YC-FHA-CN-JPH-LC:22} and LiDAR~\cite{CN-JGR-HIC:14};
and further for handling navigation pipelines~\cite{GW:17,SK-AA-NA:21} 
to efficiently and safely reach an
AoI. We use the popular frontier-based exploration~\cite{BY:98}, which
has been successfully used to explore challenging
environments~\cite{ZS:20}.

AoIs are evaluated typically on the potential quality of information.
Often, Shannon entropy~\cite{CES:48} is used for this measure of uncertainty.
In~\cite{AS-CN-SM:24-ral}, Behavioral Entropy (BE), a generalized
entropy measure,
was proposed to subjectively measure uncertainty using models of
uncertainty perception~\cite{DP:98} from the Behavioral Economics
literature. It was shown in~\cite{AS-CN-SM:24-ral, WAS-AS-CNG:25} 
that
BE provides a more effective, and flexible way to evaluate
uncertainty, especially in exploration objectives, leading to faster
and better coverage compared to standard entropy. Moreover, being
human-inspired makes it compatible with other mapping schemes that target
high-entropy areas first; then move on to
other areas.
  
In order to efficiently assign AoIs among robots, we look towards the
literature on task allocation~\cite{AS-SLS:17}. 
Both centralized solutions~\cite{HWK:55} and distributed
solutions~\cite{SC-GN-MR-ME:17} are known to be computationally and
memory-intensive for large problems. Moreover, they are mostly
addressed in static scenarios. Often authors~\cite{NR-SSK:23}
restrict the type of functions in order to develop a
tractable solution to the NP-hard problem. Game-theoretic models have
also been proposed to find solutions to task allocation problems,
where each agent is equipped with an appropriate utility
function~\cite{RK-RC-DG-JRM:22}, 
and the
optimal task allocation is related to the Nash equilibrium of this
game. Any Nash-seeking~\cite{PF-MK-TB:10} algorithm returns a
solution; but often, these algorithms require strong assumptions on
the utility functions and their derivatives. Often, complete and
perfect information on agents' utilities is required to run the
algorithm updates, while in practice only imperfect information about
tasks and other agents' capabilities is available. Further, to address the lack
of information, each agent applies consensus~\cite{MY-GH:17}, 
or gossip-based algorithms~\cite{FS-LP:16} to
estimate all other agents' strategies and compute the gradient of its
own utility. There is still a need for scalable and distributed
solutions that dynamically adapt to rewards revealed online with
performance guarantees. This is studied in~\cite{NM-MK-SM:25-tac}, but
for deterministic reward sequences. Here, we deal with more realistic
stochastic approximation sequences.

\paragraph*{Contributions}
We provide a novel distributed task allocation algorithm based on a
game-theoretic interpretation of the classical task allocation
problem.  
  The utility of the robots is given by the expected value of a random
  variable with an unknown distribution. Compared to other works, we
  give probabilistic convergence guarantees for band estimations of
  this utility. We also give out-of-sample guarantees when the utility
  is approximated using i.i.d. samples of the random
  variable. Further, we extend the idea of BE to a team of
  agents. This allows us study a large variety of heterogeneous and
  homogeneous teams and give provable justification of the benefits of
  heterogeneity for exploration. We simulate various teams of robots
  mapping a complex environment and gather performance results based
  on time taken and path-length traveled. This allows us to provide
  heuristic guidelines for choosing the team composition in terms of
  behaviors for a multi-agent exploration team.

\section{Preliminaries} \label{sec:prelim} Here, we introduce 
well-known concepts that are used throughout the paper.\footnote{
The sets of real numbers, non-negative real numbers, and non-negative
integers are denoted as $\real$, $\real_{\geq 0}$, and $\intpos$,
respectively. For a set $\set$, we denote by $|\set|$ its cardinality,
$2^\set$ its power set, by $\set^n$ its $n$ Cartesian product, and by
$\set^{n \times m}$ is the collection of $n \times m$ matrices whose
$(i,j)\tth$ entry lies in $\set$. For $x \in \real$, $[x]_0^1 \ldef \max \{0, \min \{ x , 1 \} \}$. Given $\set$, we define
$\submax \set \ldef \max \{s \in \set \,|\, s \neq \max
\,\set\}$. 
The probability of an event (over a measurable space, which should be
clear from the context) is 
$\Pr \{\cdot \}$.  }

\paragraph{Game theoretic notions} 
A \emph{strategic form game}~\cite{YN:14} is a tuple
$\game \ldef \left\langle \agt,\{\S_i\}_{i\in \agt}, \{\psi_i\}_{i \in
    \agt} \right\rangle$ consisting of a set of \emph{players} (or
\emph{agents} or \emph{robots}) $\agt$; a set of \emph{strategies}
$s_i \in \S_i$ available to each $i \in \agt$; and a set of
\emph{utility functions} $\psi_i : \times_{i \in \agt}\S_i \to \real$
over the strategy profiles of all the agents.
Denote $s_{-i}$ as the strategy profile of all
players other than $i \in \agt$. This helps to define the Nash equilibria next.

\begin{define}[Nash equilibrium]
\label{def:ne}
The strategy profile $(\hat{s}_i,\hat{s}_{-i})$ is a Nash equilibrium
(NE) of $\game$ if and only if
\begin{align*}
  \psi_i(\hat{s}_i,\hat{s}_{-i}) \geq \psi_i(s_i,\hat{s}_{-i}),
  \,\, \forall s_i \in \S_i, \,\, \forall i \in \agt\,. 
\end{align*}
$\ne(\game)$ denotes the set of all Nash equilibria of $\game$.
\bulletend
\end{define}

\paragraph{Graph theoretic notions}
A \emph{directed graph} $\grph \ldef (\agt,\edg)$, is a tuple
consisting of a set of \emph{nodes} (here robots, $\agt$),
and a set of \emph{arcs} $\edg \subseteq \agt \times \agt$ between the
nodes; see~\cite{RD:17}.  The set
$\neigh_i \ldef \{ j \in \agt \,|\, (j,i) \in \edg\}$  denotes the
set of (in) \emph{neighbors} of agent $i \in \agt$ and
$\neighb_i \ldef \neigh_i \cup \{i\}$. A \emph{path} in $\grph$ is an
ordered set of non-repeated nodes such that each pair of adjacent
nodes defines an edge. The graph $\grph$ is said to be \emph{strongly
  connected} if there exists a path from every node to every other
node. The \emph{diameter} of the graph $\diam(\grph)$ is the length of
the largest possible path between any two nodes.

\paragraph{Concentration of measure} 
Suppose that
$\xi \in \real^m$ is a random variable with an unknown probability
distribution, $\distp^*$. Assume that
$\{\hat{\xi}(k) \in \real^m \}_{k = 1}^N$ are i.i.d samples generated
by this unknown distribution, $\distp^*$. In addition, suppose that
$\distp^*$ is supported on $\Xi \subseteq \real^m$ and
$\exists \, a > 1$ such that
$\expect[\xi \sim \bbp^*]{\exp(\|\xi\|^a)} < \infty$
(i.e.~$\distp^*$ is light-tailed).  Let $\mathcal{L}(\varXi)$
be the set of all probability distributions $\mathbb{Q}$ on $\Xi$ with
bounded first moment.  Then, the $1$-Wasserstein distance between two
distributions $\bbq_1, \bbq_2 \in \mathcal{L}(\Xi)$ is defined as
\begin{align*}
  & d_\mathsf{W}(\bbq_1, \bbq_2)
    \ldef \inf_{\pi \in \Pi(\bbq_1,\bbq_2)} \int_{\Xi^2}
    \|\xi_1 - \xi_2\|_1 \, \pi(\dg\xi_1, \dg\xi_2)\,, 
\end{align*}
where, $\Pi(\bbq_1,\bbq_2)$ is the set of joint probability
distributions of $\xi_1$ and $\xi_2$ with marginals $\bbq_1$ and
$\bbq_2$ respectively. Further,
\begin{align*}
	\mathcal{B}_\varepsilon (\distp)
\ldef \{ \bbq \in \mathcal{L}(\Xi) \,|\, d_\mathsf{W}(\bbq, \distp) < \varepsilon \},
\end{align*}
is the Wasserstein ball of radius $\varepsilon$ around $\distp \in \mathcal{L}(\Xi)$.

The following result characterizes the closeness of $\bbp^*$ to the
sample average distribution
$\distphat_{N} \ldef \frac{1}{N} \sum_{k = 1}^{N}
\delta_{\hat{\xi}(k)}$. This is useful in distributional robust optimization~\cite{PME-DK:17}.
\begin{theorem}
\cite{DB-JC-SM:19-ecc}
\label{thm:dro_bound}
Suppose $\theta \in (0,1)$, and 
\begin{align}
	\varepsilon = 
	\begin{cases}
          \left( \frac{\log(c_1 \theta^{-1})}{c_2 N} \right)^{1/\max\{m,2\}},
          & \text{if } N \geq \frac{\log(c_1 \theta^{-1})}{c_2}\,; \\
          \left( \frac{\log(c_1 \theta^{-1})}{c_2 N} \right)^{1/a},
          & \text{if } N < \frac{\log(c_1 \theta^{-1})}{c_2} \,;
	\end{cases}
	\label{eq:dro_bound}
\end{align}
where $c_1$ , $c_2$ are positive constants that only depend on $a$ and
$m$.  Then,
$ \Pr \big\{ \bbp^* \in \mathcal{B}_\varepsilon
(\distphat_{N}) \big\} \geq 1 - \theta$.
  \proofend
\end{theorem} 

\section{Problem Formulation} \label{sec:problem}

Here, we first describe the constituents of our framework and then define the problem statement.

\paragraph*{Exploration environment} A team of $n$ robots
$\agt \ldef \until{n}$ is to explore an unknown environment
$\sspace \subset \real^2$ in an efficient and distributed fashion.
The environment $\sspace$ is discretized into grid elements $\grid$
that define an occupancy map $\map$.  More precisely, $\map$ is
defined in terms of a probability function
$\focc : \grid \rightarrow [0,1]$, which assigns a probability of
occupancy to each cell in $\grid$ (\emph{i.e.} $\map =
\focc(\grid)$). 
In what follows, we assume there is a subroutine that takes care
  of map merging (either centralized or distributed) and that $\map$
  is available to all robots. Instead, we focus on the distributed
  exploration part of the approach.
  As the robots explore, they aim to reduce the volume of the unknown
  set $\{x \in \grid \,|\, \focc(x) \in (0,0.5) \cup (0.5,1)\}$ while
  increasing the volume of the known set
  $ \{x \in \grid \,|\, \focc(x) \in \{0,1\}\}$. Ideally, the team
  completes the exploration when $\grid$ 
  becomes known. In practice, limited observability due to the sensing
  noise and presence of obstacles hinders uncertainty reduction.

\paragraph*{Robot Team} Each robot $i \in \agt$ uses a Bayesian
  algorithm
  to update the occupancy map in its immediate environment as well as
  its position, $x_i \in \grid$;    
see~~\cite{ST-WB-DF:05, CN-JGR-HIC:14}.  To update the map, each robot~$i \in \agt$
  identifies a set of frontiers $\tsk_i$ from the current
  $\map$.
We note that the size of
$\tsk_i \equiv \tsk_i(t)$, $i \in\agt$, may vary with time as robots
discover frontiers to visit at time $t$ (we omit the time argument when clear from the context).  To aid in exploration,
robot $i \in \agt$
associates a reward of 
$\rho_i(q) \ldef \expect{R_i^q}$ for $q \in \tsk_i$. Here $R_i^q$ is a
random variable with an unknown distribution, the expectation of
  which is to be estimated. We give further details of this procedure
in Section~\ref{sec:dist_assignment}. Randomness may appear either due to
noisy sensor measurements or environmental uncertainty.
Note that as these rewards are robot-dependent, they
introduce heterogeneity, which will influence exploration.

\paragraph*{Frontier assignment problem} 

Given $t \in \integernonnegative$, assume that
$\{\tsk_i(t)\}_{i \in \agt}$ is the set of discovered frontiers at
time $t$, and define $\tsk(t) \ldef \bigcup_{i \in \agt}
\tsk_i(t)$. 
The optimal frontier assignment at time $t$ is the solution to
\begin{subequations}
\begin{align}
  \label{eq:partition_reward} & \max_{{\salloc = (\alloc_1,\cdots,\alloc_n)  \subseteq \times_{i=1}^n \tsk_i(t)}}
                                \quad J(\salloc)\ldef \sum_{i \in \agt} \sum_{q \in \alloc_i} \rho_i(q) 
  \\
  \label{eq:partition_constraint} & \mathrm{s.t.}
                                    \quad \bigcup_{i \in \agt} \alloc_i = \tsk(t);
                                    \quad \alloc_i \cap \alloc_j = \varnothing,\,\, \mathrm{if}\,\, i \neq j.
\end{align}
\label{eq:partition_optimization}
\end{subequations}
Here, $\alloc_i \subseteq \tsk_i(t)$ is the set of all frontiers
assigned to robot $i \in \agt$ and
$\salloc = (\alloc_1,\cdots,\alloc_n)$ is the ordered collection of
sets that defines a partition of $\tsk (t)$ (as
in~\eqref{eq:partition_constraint}). The set
$\opt(\agt,\tsk,\{\rho_i(q)\}_{q \in \tsk,i\in \agt})$ collects the
optimizers of~\eqref{eq:partition_optimization}.  To ensure
  scalability to large environments, robots can only choose reachable
  frontiers within region $\tsk_i \subseteq \mathcal{Q}_i$,
around each robot's location.  
However, this might result into a
robot $i \in \agt$ running out of frontiers (\emph{i.e.}
$\tsk_i = \varnothing$). When this happens, we allow the robot to
expand its footprint radius until some frontiers become available
(\emph{i.e.}  $\tsk_i \neq \varnothing$) to explore. Thus, this
adaptive technique trades off manageable computations with potential ineffectiveness.

After assignment, each robot keeps a list
$\mathcal{T}_i$ of frontiers (a task buffer) that it is scheduled to
visit.
We assume that the task-assignment problem is
re-solved after a robot completes the tasks in its buffer (\emph{i.e.}
$\mathcal{T}_i = \varnothing$). 
Thus, triggered by one robot at time $t_e$, all other robots
will share with neighbors in a communication
  graph 
the frontiers that they have not been visited yet, as well as the new
ones they have discovered, entering a frontier re-assignment
  phase.  To do this, we will assume that the
  robot-communication graph becomes connected over time; see
  Assumption~\ref{asmp:comm_net}, so there is a time $t>t_e$ when the
  assignment problem~\eqref{eq:partition_optimization} is resolved.  The robot with an empty buffer
then updates $\mathcal{T}_i$ using
$\opt(\agt,\tsk(t_e),\{\rho_i(q)\}_{q \in \tsk,i\in \agt})$.
We will assume that the discovery of new frontiers happens at a
sufficiently slow pace for the fast re-computation of the solution
to~\eqref{eq:partition_optimization}. 
We focus on understanding the benefits of exploration
with heterogeneous rewards, leaving the question of efficient dynamic
task assignment under partial information for future work.

\paragraph*{Communication network} The robots communicate with each
other over a time-varying graph $\grph(t) \ldef (\agt,\edg(t))$. The
arc set $\edg(t)$ defines the connections between robots at time
$t \in \integernonnegative$: in this way, $(i,j) \in \edg(t)$ if and
only if robot $i \in \agt$ can send information to robot $j \in \agt$
at time $t$, which can only occur when they are within a distance of
each other. Further we assume the following.

\begin{assume} \label{asmp:comm_net} \thmtitle{Periodic fast
    communication} For the communication network sequence
  $\{\grph(t) = (\agt,\edg(t))\}_{t \in \integernonnegative}$, there exists a (known)
  $\tau \in \integernonnegative$ such that the graph
  $\bar{\grph} \ldef (\agt,\bar{\edg}(k))$, 
 with $\bar{\edg}(k) \ldef \bigcup_{t = 1}^\tau \edg(k\tau + t)$, is strongly
  connected $\forall k \in \integernonnegative$. 
  Further, robot communication happens
  at a faster time scale than robot movement.   
  \bulletend
\end{assume}

\paragraph*{Problem Statement} 
Given the aforementioned setup,
\begin{enumerate}
\item Provide a robust and 
  distributed frontier allocation algorithm that converges to an
  stochastic optimal frontier assignment under time-varying
  communications.
  
\item Study the benefits of BE heterogeneity on exploration.

\item Verify the performance of the algorithms in complex, high-fidelity simulation
  environments and provide heuristic guidelines on choosing heterogeneous
  parameters in a mixed group autonomous agents, possibly including
  humans.

\end{enumerate}

\section{Distributed Frontier Assignment} \label{sec:dist_assignment}

Here, we first tackle the problem of efficient frontier
allocation. This works for any evaluation scheme of frontiers. We
provide the specific frontier utility metric in
Section~\ref{sec:behavioral_entropy}.

In order to assign frontiers in a distributed fashion, we generalize
the distributed projected best-response ascending gradient (d-PBRAG) dynamics
proposed
in~\cite{NM-MK-SM:25-tac}. 
The following definition from~\cite{NM-MK-SM:25-tac} helps us in
characterizing the optimizers of~\eqref{eq:partition_optimization},
$\opt(t)$ at a particular $t \in \integernonnegative$.
\begin{define} \label{def:dominating_agent} 
\thmtitle{Frontier specific dominating agent} 
A robot $i \in \agt$ is said to \emph{dominate frontier $q \in \tsk(t)$}, if
$q \in \tsk_i(t)$ and
$\rho_i(q) \geq \rho_j(q)$, $\forall j \in \agt$ such that
$q \in \tsk_j(t)$.  If a frontier $q \in \tsk$ has exactly one
dominating robot, we say that there exists a \emph{unique dominating
  agent for $q$}.  \bulletend
\end{define}

In~\cite{NM-MK-SM:25-tac} it was shown that if there is a unique
dominating agent per $q \in \tsk(t) = \bigcup_{i \in \agt} \tsk_i(t)$,
then $\opt(t)$ is in fact a singleton set. Moreover, it is also the
unique Nash equilibrium of the game
\begin{align}
	\game \ldef {\left<\agt,\{\w_i\}_{i \in \agt}, \{U_i\}_{i \in \agt} \right>},
	\label{eq:game}
\end{align}
where robot $i \in \agt$ is equipped with the utility function,
\begin{align}
  U_i(\w_i,\w_{-i}) & = \hspace{-1ex}
                      \sum_{q \in \tsk_i(t)} \bigg[ \rho_i(q)w^q_i - \hspace{-1ex}
                      \max_{\substack{j \in \agt \setminus \{i\} \\ q \in \tsk_j}} \rho_j(q)w^q_j w^q_i \bigg] ,
	\label{eq:agent_reward_max} 
\end{align}
and the strategies available to robot $i \in \agt$ are
$(w_i^q \in [0,1])_{q \in \tsk_i(t)}$, with
$\w_i = (w_i^q)_{q \in \tsk_i(t)}$.
A weight $w_i^q = 1$ (resp.~$w_i^q = 0$), loosely means that agent
$i \in \agt$ is assigned (resp.~not assigned) the frontier
$q \in \tsk_i(t)$. Note that sufficiently small and constant
perturbations to individual rewards can break down ties in allocation
and create unique dominating agents per task, aiding in the frontier
allocation problem. In this work, we show in
Section~\ref{sec:behavioral_entropy} how agent heterogeneity can
naturally encode this feature.

Recall that $\rho_i(q) = \expect{R_i^q}$ is the reward that robot
$i \in \agt$ gets for $q \in \tsk_i$. We assume that the random
variables $\{R_i^q\}_{i \in \agt, q \in \tsk}$ are independent of each
other for all $i,q$.  In most real world scenarios, robots do not have
access to the actual distribution of rewards, and just receive samples
of them.
Using these, we assume that the robot can approximate its reward with
some sequence $\{z_i^q(t)\}_{t \in \integernonnegative}$ (e.g.~a
  sampled average value). With this in mind, we define the following.
\begin{define}
  \thmtitle{Probabilistic band convergence} 
  \label{def:conv}
  We say that a sequence of random variables
  $\{z_i^q(t)\}_{t \in \integernonnegative}$ \emph{converges in
    probability to a $\mu$-band} around $\rho_i(q) \in \real$, with
  $\mu \geq 0$, if $\forall \epsilon>0$,
  $\prob{|z_i^q(t) - \rho_i(q)| - \mu > \epsilon} \to 0$ as
  $t \to \infty$.
\bulletend
\end{define}
Note that if the reward $\rho_i(q)$ is some other function (and not
the expected value) of the random variable $R_i^q$, then the previous
definition is in line with the standard notion of concentration
estimation of $\rho_i(q)$. Since we are interested in characterizing
the convergence w.r.t.~the sample size, we choose the expected
value 
as the reward.

Recall that the robots communicate over a time-varying graph
$\grph(t) = (\agt,\edg(t))$. This allows us to allocate the
frontiers in a distributed manner while providing an algorithm that
seeks the NE of $\game$ in~\eqref{eq:game}, and accounting for
increasingly more accurate approximations
$\{z_i^q(t)\}_{t \in \integernonnegative}$ of $\rho_i(q)$, for
$i \in \agt$. Thus, to compute their utility gradients and update
strategies $w^q_i$ simultaneously, the robots make use of the
following dynamics given in~\cite{NM-MK-SM:25-tac}:
\begin{subequations}
\begin{align}
  \label{eq:dist_wt_update} & w^q_i(t+1) \hspace{-0.5ex} = \hspace{-0.5ex}
                              \Big[ w^q_i(t)
                              + \gamma^q_i (t) \Big(\fseq^q_i(t) -
                              \frac{1}{2} \big( M^q_i(t) + S^q_i(t) \big) \Big)   \Big]_0^1 , \\
  \label{eq:max_consensus} & M^q_i(t+1) \hspace{-0.5ex} = 
                             \hspace{-0.5ex} \switch
                             \Big(\max_{j \in \neighb_i} M^q_j(t), e^q_i(t+1), t+1, T \Big),\\
  \notag & S^q_i(t+1) = \switch \Big(\submax
           \Big\{\{S^q_j(t)\}_{j \in \neighb_i},M^q_i(t),e^q_i(t) \Big\}, \\
  \label{eq:submax_consensus} & \hspace{12em} e^q_i(t+1), t+1, T \Big),\\
  \label{eq:input_injection} & e^q_i(t+1) =
                               \switch \Big(e^q_i(t),z^q_i(t+1),t+1,T \Big)\,,
\end{align}
\label{eq:dist_pbrag}
\end{subequations}
for some $T \in \realnonnegative$, and
where $\switch$ is the switching function
\begin{align}
	\switch \Big(m, \fseq,  t, T  \Big) \ldef 
	\begin{cases}
          \fseq, &\mathrm{if}\,\, t \mod T = 0,\\
          m, &\mathrm{otherwise}\,.
	\end{cases}
	\label{eq:switch}
\end{align}
The previous dynamics combines a max consensus sub-routine with a
periodic knowledge injection to steer the strategies of each agent
towards the NE of $\game$. 

Work in~\cite{NM-MK-SM:25-tac} considered deterministic
  rewards and a deterministic sequence converging to the true value of
  the rewards. It was shown that that by choosing the step size
  $\gamma_i^q(t)$ in a certain way, it is possible to send the weight
  $w_{i^*_q}^q(t)$ of the dominating agent ($i^*_q$ for
  $q \in \tsk_i$) to \emph{one} while simultaneously reducing
  $w_j^q(t)$ (for $j \neq i^*_q$) to \emph{zero}. This stems from the
  fact that the max consensus subroutine
  in~\eqref{eq:max_consensus},~\eqref{eq:submax_consensus} stabilizes
  within a number of time steps dependent on the diameter of the
  communication graph and then the average
  $0.5 \big(M^q_i(t) + S^q_i(t)\big)$ increases (or decreases) the
  weights $w_i^q(t)$ appropriately. 
Going beyond the analysis provided in~\cite{NM-MK-SM:25-tac}, we show next
that this algorithm converges in probability to a solution of the task
assignment problem~\eqref{eq:partition_optimization} for a certain
class of stochastic approximation sequences. To keep the notations
clean, we present the arguments with $\tsk_i(t) = \tsk$,
$\forall\, i \in \agt$ (\emph{i.e.} every robot
has access to every frontier). The algorithm can be easily adapted to
the partial frontier setup that this paper deals with. We refer to Appendix~\ref{sec:proofs} for proofs.

\begin{theorem}
  \thmtitle{d-PBRAG convergence in
      probability}
\label{thm:pbrag_converge}
Suppose the communication graph $\grph(t)$ satisfies
Assumption~\ref{asmp:comm_net} with connectivity period
$\tau$. Suppose that each $q \in \tsk$ has a unique dominating agent and
let
\begin{align}
	\mu^q < 0.5 \, (\max \{\rho_i^q\}_{i \in \agt} - \submax
\{\rho_i^q\}_{i \in \agt}), \,\, \forall q \in \tsk\,.
	\label{eq:band_converge}
\end{align}
Further for each $i \in \agt$, $q \in \tsk$,
assume that $z_i^q(t)$ converges in probability to a $\mu^q$ band around
$\rho_i(q)$. 
Consider an initial condition
$(\w_i(0) \in [0,1]^{|\tsk_i|})_{i \in \agt}$, let
$(\w_i(t))_{i \in \agt}$ be the solution to~\eqref{eq:dist_pbrag} with
$T > 2\,\tau + 1$, and
\begin{align*}
	\gamma^q_i(t) = 
	\begin{cases}
		\alpha^q_i(k) \geq 0, & \mathrm{if}\, t \in \{kT,\cdots,kT+2\tau-1\},\\
		\beta^q_i(k)>0, & \mathrm{if}\, t \in \{kT+2\tau,\cdots,kT+T-1\},
	\end{cases}
\end{align*}
$\forall i \in \agt$, $\forall q \in \tsk$, with $k \in
\intpos$. Further, for all $ i \in \agt$, $\forall q \in \tsk$; take
sequences $\alpha^q_i(k) \to 0$ as $k \to \infty$ and
$\beta^q_i(k) \to \infty$ as $k \to \infty$. Then, for each
$ i \in \agt$, and $\varepsilon > 0$,
\begin{align}
  \prob{\|\w_i(t) - \wb_i\| > \epsilon} \to 0, \qquad \mathrm{as}\,\, t \to \infty,
\end{align}
where
$(\{q \in \tsk_i \,|\, \bar{w}^q_i = 1\})_{i \in \agt} \subseteq
\opt(\agt,\tsk,\{\rho_i(q)\}_{q \in \tsk,i\in \agt})$ is a solution
to~\eqref{eq:partition_optimization}, 
and $\wb_i \ldef (w_i^q)_{q \in \tsk_i}$.
\proofend
\end{theorem}

Next, we provide further robustness guarantees
for a particular estimator of the expected rewards.  
Let us denote the samples of the random variable $R_i^q$ as
$\{\hat{R}_i^q(k)\}_{k \in \{1, \cdots, N_i\}}$. 
For each robot $i \in \agt$ and each frontier $q \in \tsk_i$, let the
sample average distribution be
$\hat{\bbp}_{N_i}^{i,q} \ldef \frac{1}{N_i} \sum_{k = 1}^{N_i}
\delta_{\hat{R}_i^q(k)}$
where $\delta_{\hat{R}_i^q(k)}$ is the Dirac
delta at
$\hat{R}_i^q(k)$.  In what follows, we assume that the robots estimate
the actual reward mean via the sequence of random variables
\begin{align}
  z_i^q(t) = \frac{1}{2}
  \bigg( \sup_{\bbq \in \mathcal{B}_\varepsilon(\hat{\bbp}_{N_i^t}^{i,q})} \hspace{-1ex} \mathbb{E}_{\xi \sim \bbq}[\xi]  \hspace{1ex} + \hspace{-1ex} \inf_{\mathbb{Q} \in \mathcal{B}_\varepsilon(\hat{\bbp}_{N_i^t}^{i,q})} \hspace{-1ex} \mathbb{E}_{\xi \sim \bbq}[\xi] \bigg).
	\label{eq:dr_sample_update}
\end{align}
where the $\varepsilon$ are chosen properly. Here, $N_i^t$ represents
the number of samples available to $i \in \agt$ at time
$t \in \integernonnegative$ and recall that
  $\mathcal{B}_\varepsilon (\hat{\bbp}_{N_i^t}^{i,q})$ 
  is the appropriate Wasserstein ball.   
Thus, the robot estimates its reward for a task as the
  average of the worst case expectations (in either direction)
  generated by the probability distributions that are at most
  $\varepsilon$ distance from its sample average. Further, if $N_i^t$
  increases, the robot can get better estimates using a larger sample
size.  In the next result (proof of which is in Appendix~\ref{sec:proofs}), we give distributionally robust
guarantees on the convergence of the d-PBRAG algorithm.

\begin{theorem}
  \thmtitle{Finite sample guarantees with distributionally robust
    d-PBRAG}
\label{thm:dr_dpbrag_convergence}
Suppose $\forall i \in \agt$,
$N_i^t \to K_i \in \integernonnegative$. 
Moreover, suppose $\theta \in (0,1)$ is chosen such that with
$\varepsilon$ in~\eqref{eq:dro_bound}, we have that, for each
$q \in \tsk$, $\exists \, i^*_q \in  \agt$ such that
\begin{align}
  \sup_{\bbq \in \mathcal{B}_\varepsilon(\hat{\bbp}_{N_j^t}^{j,q})}
  \hspace{-1ex}\mathbb{E}_{\xi \sim \bbq}[\xi] \hspace{1ex} < \inf_{\bbq \in \mathcal{B}_\varepsilon(\hat{\bbp}_{N_{i^*_q}^t}^{{i^*_q},q})} \hspace{-1ex} \mathbb{E}_{\xi \sim \bbq}[\xi], \hspace{1ex}\forall j \neq i^*_q\,.
  \label{eq:finite_bound}
\end{align} 
Finally, consider an initial condition
$(\w_i(0) \in [0,1]^{|\tsk_i|})_{i \in \agt}$ and let
$(\w_i(t))_{i \in \agt}$ be the solution to~\eqref{eq:dist_pbrag} with
$T$ and $\gamma_i^q(t)$ as in Theorem~\ref{thm:pbrag_converge}, and
$z_i^q(t)$ as in~\eqref{eq:dr_sample_update}.  Then, with probability
at least $(1- \theta)$, $(\w_i(t))_{i \in \agt}$ converge in
  probability to a 0 band around
$(\wb_i)_{i \in \agt}$; where,
$(\{q \in \tsk_i \,|\, \bar{w}^q_i = 1\})_{i \in \agt} \subseteq
\opt(\agt,\tsk,\{\expect{\xi_i^q}\}_{q \in \tsk,i\in \agt})$ is a
solution
to~\eqref{eq:partition_optimization}. 
\proofend
\end{theorem} 

The previous result states that when the robots only have access to
finite samples from the original distribution, then it is possible to
give a confidence interval, using $\theta$, with which the converging
strategies of the game $\game$ is an optimal frontier partition. We
end this section by saying that if $N_i^t \to \infty$ in
Theorem~\ref{thm:dr_dpbrag_convergence}, then $\theta$ can be made
arbitrary close to \emph{zero}.  


\section{Heterogeneous Behavioral
  Exploration} \label{sec:behavioral_entropy} Here, we first introduce
the notion of generalized BE~\cite{AS-CN-SM:24-ral} to
determine a system of weighted rewards per robot, $\rho_i$,
$i \in \agt$. Later we show the benefits of such heterogeneous
behavior for exploration.

Given a discretization $\grid$ of the environment, let $p_k \in [0,1]$
(resp. $(1-p_k) \in [0,1]$) be the probability that the cell
$k \in \grid$ is a free space (resp.~an obstacle). 
Each robot perceives this uncertainty differently, and we use
\emph{probability weighting functions} to further encode this
heterogeneity.  
To do this, we employ the popular Prelec's weighing
function~\cite{DP:98}, $\omega : [0,1] \rightarrow [0,1]$,
\begin{equation}
\label{eq:prelec}
\omega(p)=e^{-\beta(-\log p)^{\alpha}}, \quad 
\alpha>0, \; \beta>0, \; \omega(0)=0\,,
\end{equation}
which transforms a probability vector $[p, 1-p]^\top$ into a
perceived probability vector $[\omega(p), \omega(1-p)]^\top$.

In order to evaluate the utility of visiting a frontier $q \in \grid$,
each robot will use the notion of BE 
$H_\alpha$~\cite{AS-CN-SM:24-ral}.
This expands the notion of Shannon's entropy 
by using Prelec's weights $\omega$
from~\eqref{eq:prelec} as
\begin{equation}
\label{eq:entropies_perceived}
H_\alpha(p) \ldef - \omega(p) \log(\omega(p)) - \omega(1-p) \log(\omega(1-p))\,.
\end{equation}
Moreover,
from~\cite{AS-CN-SM:24-ral}, setting
$\beta=\exp((1-\alpha)\log(\log(2)))$
guarantees satisfaction of entropy axioms~\cite{CES:48,JMA-SGB-SH:18}.
In~\cite{AS-CN-SM:24-ral}, 
it was evaluated how low $\alpha$  
model an ``uncertainty averse'' 
behavior in exploration,
where the robot has a lower threshold for uncertainty.
With high $\alpha$
values we get ``uncertainty
insensitive'' behavior, implying that the robot considers more certain
outcomes.
The robot computes this entropy per cell in its sensing region around the frontier $q \in \grid$ to get the utility of $q$. We give more details next.

\paragraph*{Information gain and utility function}
A sensor footprint $\mathcal{Q} \subset \grid$ is
obtained around each frontier $q \in  \mathcal{F}_i(t)$. This is a
collection of occupancy cells according to the sensor model. The
robots can use any standard LiDAR model~\cite{ST-WB-DF:05} to update
the costmap in their sensing area.
Suppose a robot $i \in \agt$ has a behavioral parameter
$\alpha_i \in \real_{>0}$.  Moreover, let $x_i(t) \in \grid$ be the
position of the robot at time $t \in \integernonnegative$ and let
$\eta(x,y)$ denote the path length between two points $x,y \in
\grid$. Then for a frontier $q \in  \tsk_i(t)$, 
the robot associates
\begin{align}
  \hspace{-1ex} \rf_i(q; \mathcal{M}) \ldef
  \sum_{k \in \mathcal{Q}} H_{\alpha_i}(p_{k}),
  \,\, \phi_i(q; \mathcal{M}) \ldef \frac{1}{\eta(x_i(t),q)}
\label{eq:beh_utility}
\end{align}
and
$R_i^q = \rf_i(q; \mathcal{M}) \phi_i(q; \mathcal{M})$
Note that the $\rf_i$ is the behavioral information gain
which is calculated as the reduction in entropy from observing the
area $\mathcal{Q}$ around the frontier $q \in  \tsk_i(t)$. Further,
$\phi_i(q; \mathcal{M})$ makes it so that the robots find closer
frontiers more lucrative than the farther ones.

Recall the robot relies on its sensors and particle filter
  like models approximate the actual of occupancy ($p_k$) of a cell
  $k \in \grid$ based on generating samples $\hat{p}_k$ of this
  occupancy value. Thus, the robot associates
  $\rho_i(q) = \expect{R_i^q} = \expect{\rf_i(q; \mathcal{M})
    \phi_i(q; \mathcal{M})}$ as the reward for visiting the
  frontier. Moreover, the robot can only calculate a sample average
  approximation of $\rf_i$.
As such, this ties in with the uncertain reward
scenario described in Section~\ref{sec:dist_assignment}. 
In what follows, to simplify the presentation, we assume that
the robots have computed a good enough approximation of the frontier
reward, $\rho_i(q)$, and focus our attention on the
effect of heterogeneity induced by the behavioral parameters.

\begin{figure}
\begin{center}
	\includegraphics[width=0.9\linewidth]{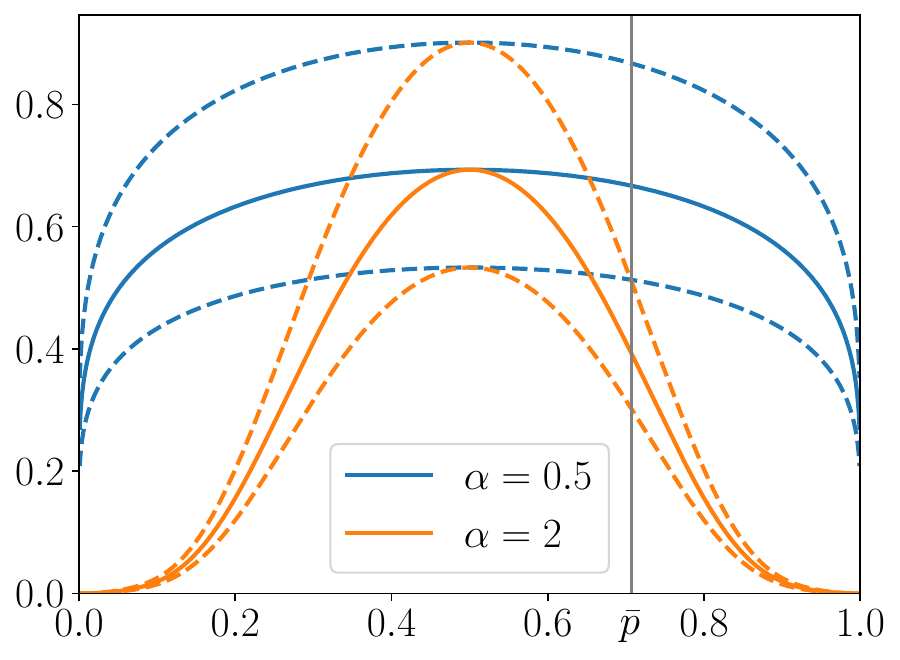}
	\caption{Scaled entropy vs. probability, \emph{i.e.}
          $c . H_\alpha(p)$ vs. $p$. Upper (resp. lower) dotted lines have $c = d_M$ (resp. $c = d_m$). Bold line has $c =1$.}
	\label{fig:entropy_band}
\end{center}
\end{figure}

\paragraph*{Mapping dynamics} Suppose that $\grid = \{1, \cdots, n\}$ is
the set of finite cells that a robot can occupy. 
As each robot is only able to map an area around a frontier, then a
team of $|\agt|$ robots would only be able to map areas around
$|\agt|$ frontiers. Thus, assume that $n = |\agt|$, in the subsequent
discussion. Further, assume that that the sensor footprint
  around cell $q \in \grid$ obeys $\mathcal{Q} = \{q\}$, \emph{i.e.}
  the robot can only map the cell it visits. We omit the discussion of
  more general footprints to reduce complexity yet highlight the
  effect of
  heterogeneity.  
Then, the mapping of the unknown environment obeys an algorithm
of the form, from $x_i(0) \in \grid$
\begin{subequations}
\begin{align}
	\label{eq:het_pos_update} x_i(t+1) & = \assign_{k \in \grid} \frac{H_{\alpha_i} (p_k(t))}{\eta(x_i(t),k)}, \quad \forall i \in \agt;\\
	\label{eq:het_prob_update} p_k(t+1) & = 
	\begin{cases}
		\belief(p_k(t)), & \text{if } k \in \{x_i(t+1)\}_{i \in \agt};\\
		p_k(t), & \text{otherwise} \,.
	\end{cases}	
\end{align}
\label{eq:het_update}
\end{subequations}
Here, $\assign_{k \in \grid} V(k)$ assigns the cell $k$ with the highest value $V$ (where, $V(k)$ is a vector of values associated with each cell grid) that is not the current position. For example, one way is to use the allocation problem~\eqref{eq:partition_optimization} and ensure unique dominating agents.

In this framework, the belief update in~\eqref{eq:het_prob_update}
corresponds to the act of mapping the cell $k \in \grid$. This could
be Bayesian update or some other form of update that incorporates
noisy sensor measurements. 
Let $p_k^* \in \{0,1\}$ be the ground truth occupancy of the cell $k$ and
assume that the mapping scheme is ``good'' in
the sense of
\begin{align}
	|\belief(p_k) - p_k^*| \leq \mapconst\,|p_k - p_k^*|\,,
	\label{eq:good_mapping} 
\end{align}
for some $\mapconst \in [0,1)$. Let the total entropy of the map be $\entropy(t) \ldef \sum_{k \in \grid} H_1(p_k(t))$,
where $H_1$ is the BE with $\alpha = 1$ (\emph{i.e.}
Shannon's entropy).  This allows us to track the entropy reduction
(from a good initial guess) under the exploration algorithm in the
next result. The proof (refer to Appendix~\ref{sec:proofs}) 
involves combining concave and differentiable
properties to bound the change.

\begin{lemma}\thmtitle{Convergence of the exploration algorithm}
\label{lem:alg_conv}
Suppose $|p_k(0) - p_k^*| < 0.5$. Then, 
under~\eqref{eq:het_update} and~\eqref{eq:good_mapping},
\begin{align}
	\notag	 \entropy(t+1) - \entropy(t) & \leq
           \sum_{i \in \agt} \Big(\log(1-p_{k_i^*}(t))- \\ 
  	\label{eq:entropy_dec} & \hspace{-5ex} \log(p_{k_i^*}(t))
           \Big) \Big(\frac{1}{2} \mapconst^t - \big(p_{k_i^*}(t) - p_{k_i^*}^* \big) \Big) ,
\end{align}
where
$k_i^* \in \argmax_{k \in \grid} H_{\alpha_i} (p_k(t)) /
\eta(x_i(t),k)$. The right hand side of~\eqref{eq:entropy_dec} is a
negative quantity. 
\proofend
\end{lemma}

We finish the section by analyzing how a multi-robot system with
different $\alpha_i$ is beneficial for mapping of an unknown
area. Essentially, under~\eqref{eq:het_update}
and~\eqref{eq:good_mapping}, each robot is guaranteed to visit a
distinct location of $\grid$ and continue the mapping
process. 
This is elaborated in the remark later. 

\begin{figure*}
	\begin{center}
	\begin{tabular}{ccl}
		\scalebox{1}{
		\begin{tikzpicture}
  		\node[inner sep=0pt] (img) at (0,0) {\includegraphics[trim=58pt 66pt 118pt 69pt, clip, scale=0.6]{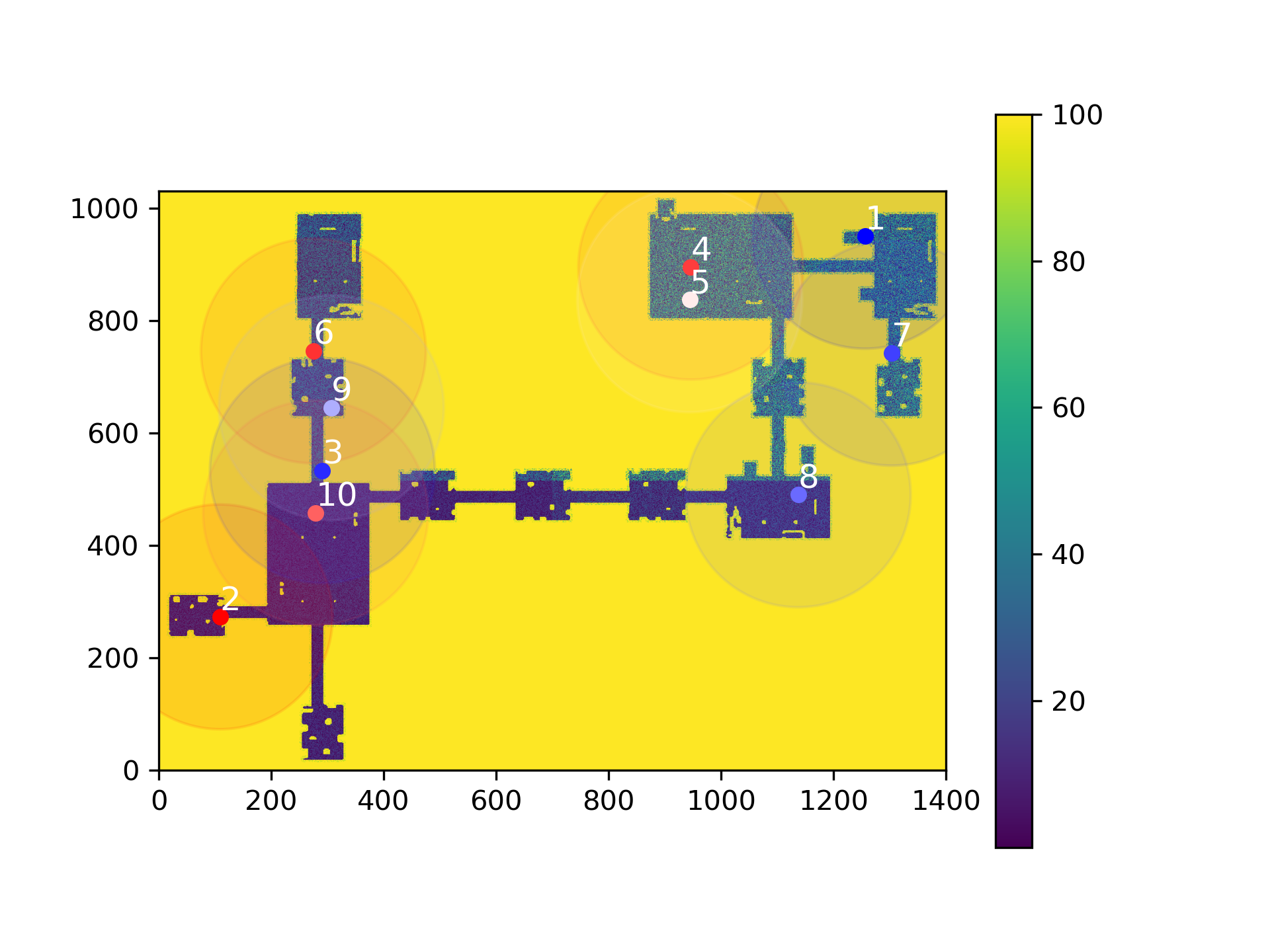}};

  		\foreach \x/\label in {0/0,1/20,2/40,3/60,4/80,5/100,6/120,7/140}
  			\pgfmathsetmacro{\xscale}{\x*0.864}
    			\draw[thick] 
      		($(img.south west) + (\xscale cm, 0)$) -- ++(0, -0.15cm)
      		node[below] {\small \label};

  		\foreach \y/\label in {0/0,1/20,2/40,3/60,4/80,5/100}
			\pgfmathsetmacro{\yscale}{\y*0.83}    
    			\draw[thick] 
      		($(img.south west) + (0, \yscale cm)$) -- ++(-0.15cm, 0)
      		node[left] {\small \label};
		\end{tikzpicture}}
 &
            \scalebox{1}{
		\begin{tikzpicture}
  		\node[inner sep=0pt] (img) at (0,0) {\includegraphics[trim=86pt 38pt 70pt 95pt, clip, scale=0.6]{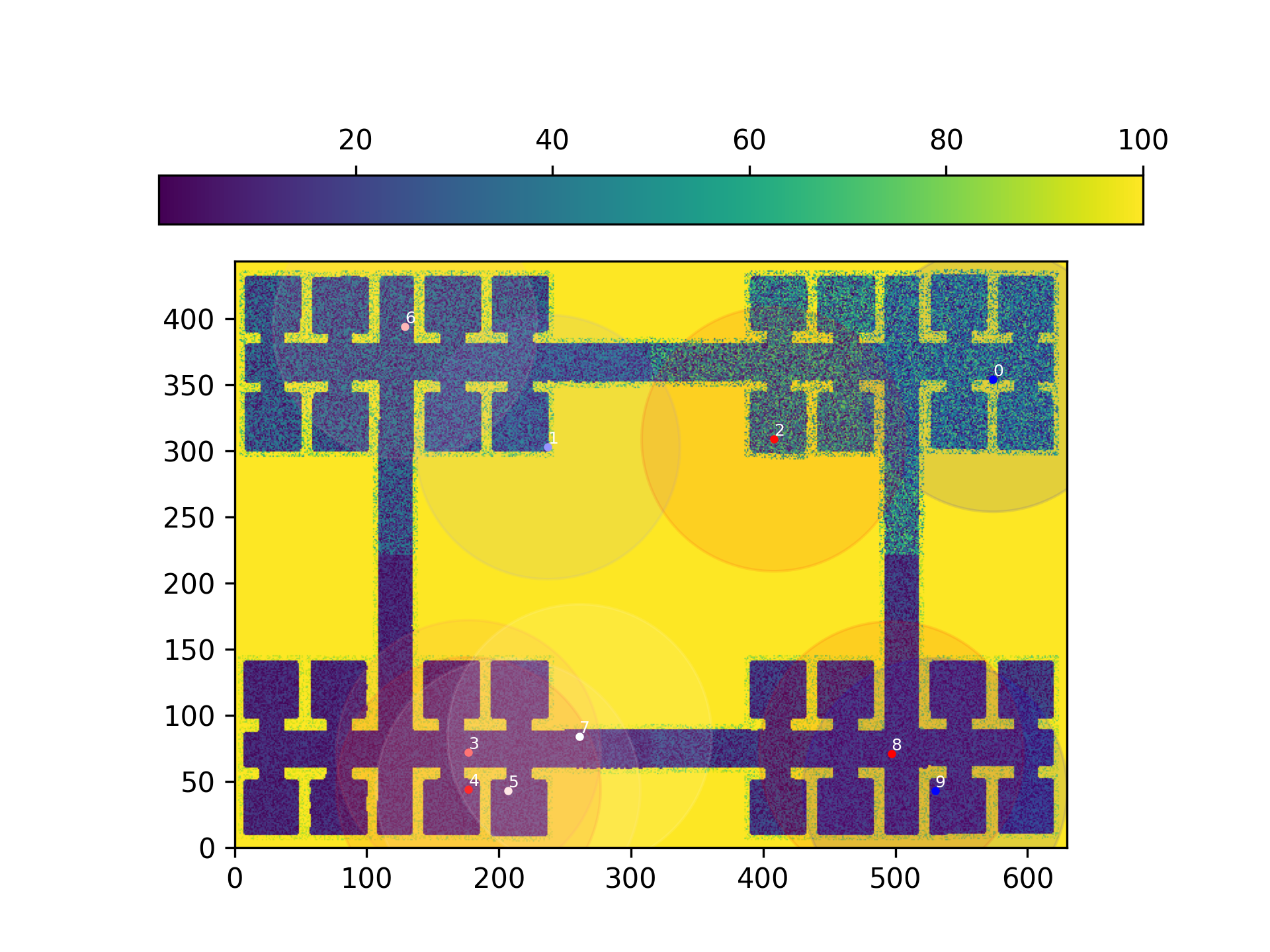}};

  		\foreach \x/\label in {0/0,1/10,2/20,3/30,4/40,5/50,6/60}
  			\pgfmathsetmacro{\xscale}{\x*1.03}
    			\draw[thick] 
      		($(img.south west) + (\xscale cm, 0)$) -- ++(0, -0.15cm)
      		node[below] {\small \label};

  		\foreach \y/\label in {0/0,1/5,2/10,3/15,4/20,5/25,6/30,7/35,8/40}
			\pgfmathsetmacro{\yscale}{\y*0.505}    
    			\draw[thick] 
      		($(img.south west) + (0, \yscale cm)$) -- ++(-0.15cm, 0)
      		node[left] {\small \label};
		\end{tikzpicture}} 
 &
		\includegraphics[trim=360pt 30pt 55pt 35pt, clip, scale=0.55]{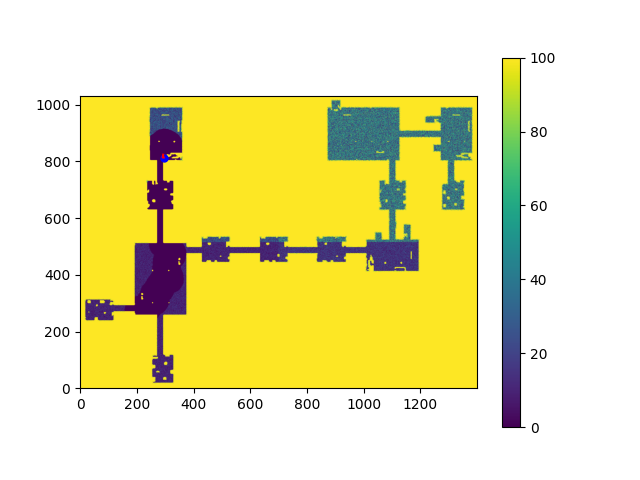}
        \\

        (a) DARPA subT section &
        (b) Urban circuit with 32 rooms
	\end{tabular}
    \caption{Maps used for simulations. The plots use the same colorbar. The robot positions $x_i(0)$ is represented by dots and the circle around each robot shows its initial region of frontier consideration. The colorbar is common across the plots.}
    \label{fig:extra_sim_environment}
    \end{center}
\end{figure*}

\begin{proposition}[On the effect of heterogeneity]
  \label{prop:het_beh} Suppose that $\{\alpha_i\}_{i \in \agt}$ is a
  set of distinct elements and define $d_M \ldef \max_{i,j,k,l \in \grid} \frac{\eta(i,j)}{\eta(k,l)}$, $d_m \ldef \min_{i,j,k,l \in \grid} \frac{\eta(i,j)}{\eta(k,l)}$.
Further, for any two $i \neq j \in \agt$, let $\bar{p}_{i,j} \ldef \min \Big( \{p \in (0,1) \,|\, d_M H_{\alpha_i}(p) = d_m H_{\alpha_j}(p) \} \cup \{p \in (0,1) \,|\, d_m H_{\alpha_i}(p) = d_M H_{\alpha_j}(p) \} \Big)$,
and define
$\bar{p} \ldef \min_{i,j \in \agt; i \neq j } \bar{p}_{i,j}$.  Suppose
that for each $ k \in \grid$,
$p_k(0) \in [0,1] \setminus [\bar{p}, 1-\bar{p}]$. Let
$(x_i(t))_{i \in \agt}$ be the solution to~\eqref{eq:het_update}
and~\eqref{eq:good_mapping} from any $x_i(0) \in \grid$, for
$ i \in \agt$. 
Then, for all $t \ge 0$, $x_i(t) \neq x_j(t)$ if
$i \neq j \in \agt$.
\proofend
\end{proposition}
The proof is described in Appendix~\ref{sec:proofs}.

\begin{remark}[Interpretation of the effect of heterogeneity]
  The range of occupancy values where a unique assignment is made
  depends on the choice of $\alpha_i$'s and the distances between the
  grid cells (see Figure~\ref{fig:entropy_band} and
  Proposition~\ref{prop:het_beh}). If $d_m$ is much smaller than
  $d_M$, then a much larger value of $\alpha_i$ is required to make
  $[\bar{p}, 1- \bar{p}]$ small. Mathematically, this is possible to
  do, but it introduces numerical issues as the entropy values become
  very small near the boundaries. As a result of
  Proposition~\ref{prop:het_beh}, a proper choice of $\alpha_i$'s
  leads to a unique cell allocation for robots and hence improve the
  efficiency of exploration.  \bulletend
\end{remark}

\section{Simulations and Results}

\begin{figure}[tbp]

    \centering
    \begin{subfigure}[t]{0.47\columnwidth}
   \centering 
  \includegraphics[trim=29pt 0pt 0pt 20pt,clip,width=\linewidth]{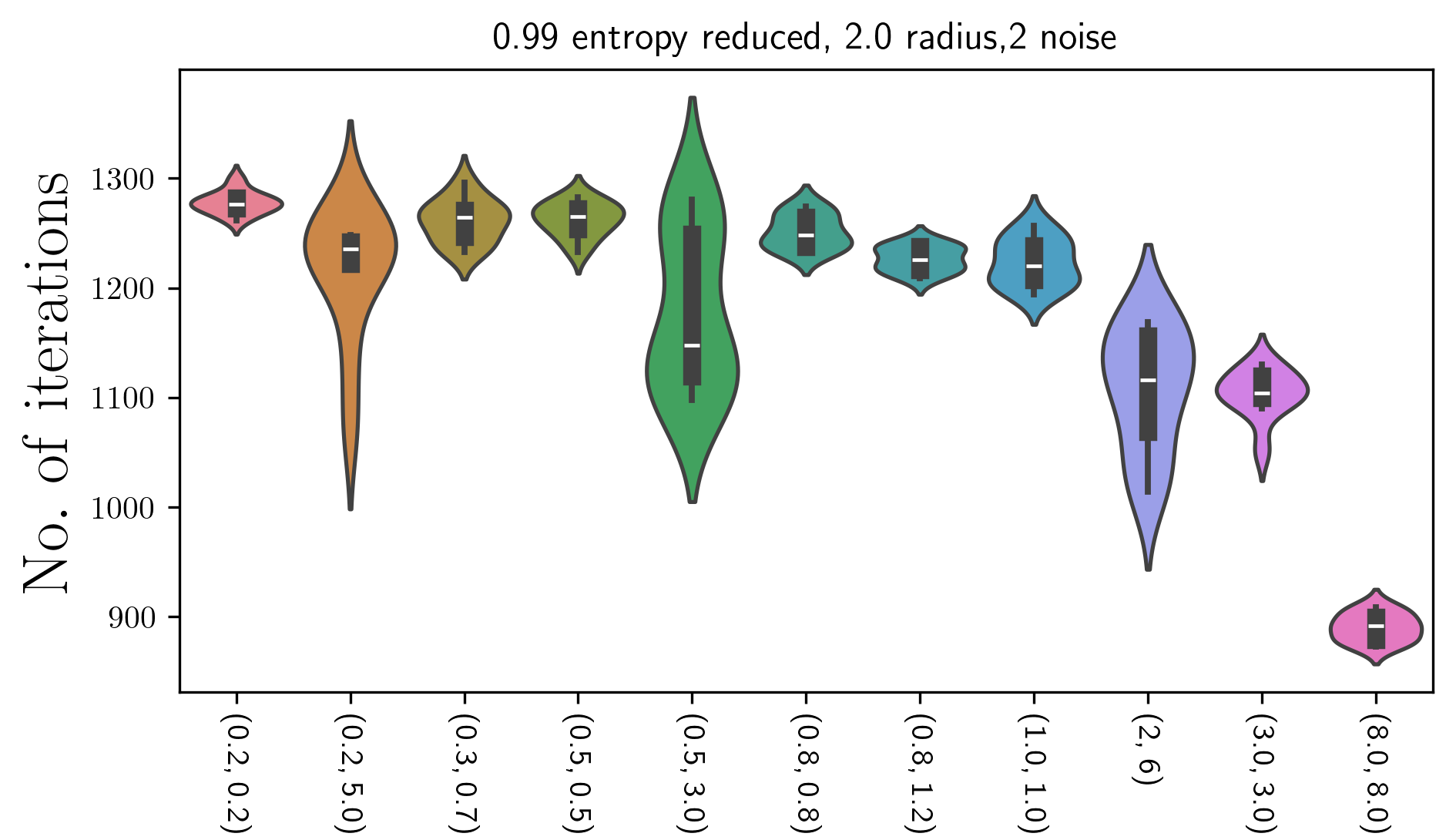}
  \caption{{\small Exploration time}}
\end{subfigure}
\medskip
\begin{subfigure}[t]{0.48\columnwidth}
   \centering 
  \includegraphics[trim=20pt 0pt 0pt 20pt,clip,width=\linewidth]{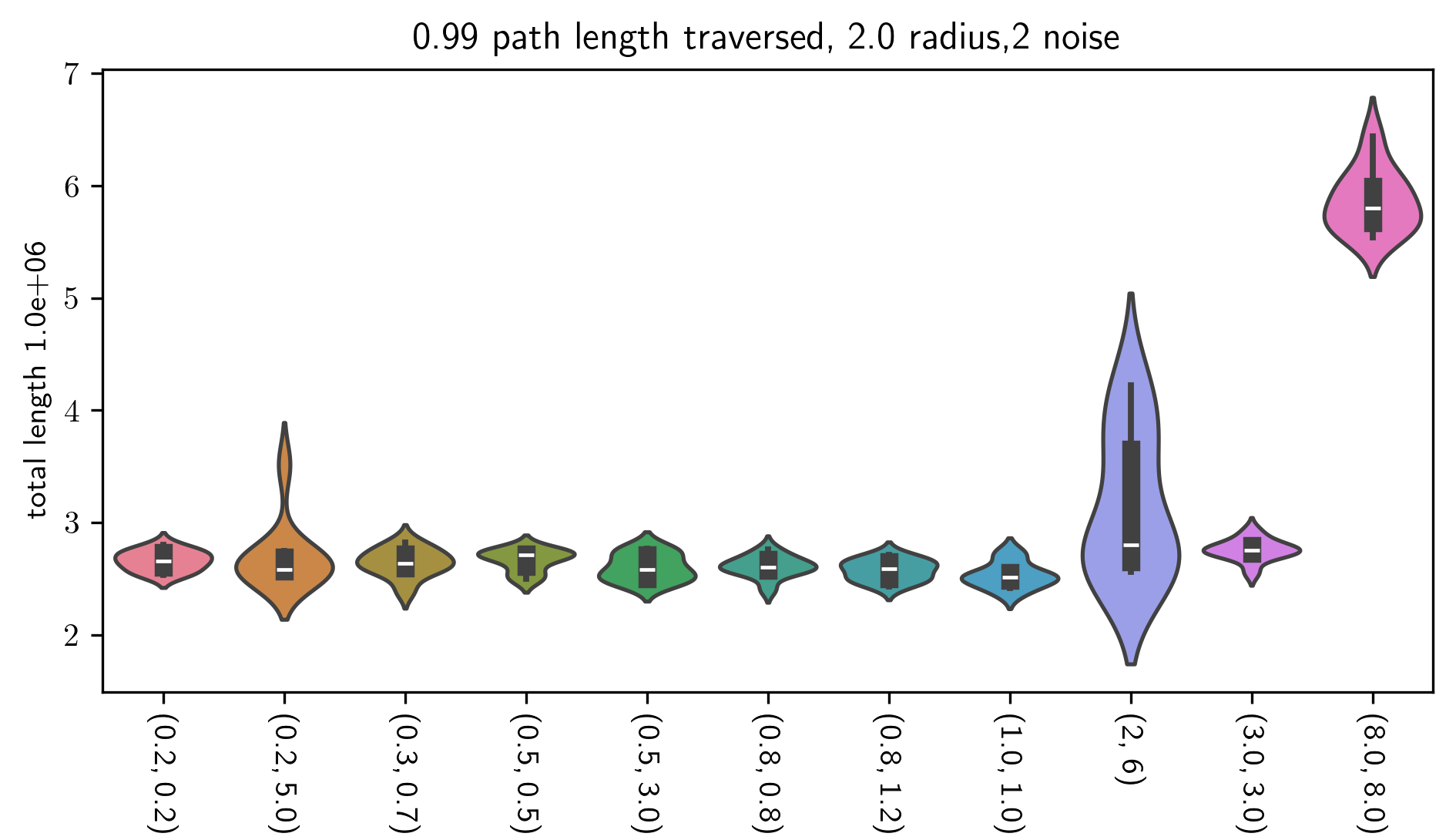}
  \caption{\small Total Path Length (in km)}
\end{subfigure}\hfil 
\vspace{-0.1in}
    \caption{\small Simulation metric for DARPA subT map, $r=2$, $\sigma^m = 2$. X-axis shows different $\alpha$ ranges $(a,b)$.
    }
    \label{fig:entropy_and_path_length}
\end{figure}

In this section, we implement the heterogeneous mapping algorithm on
Python environments. First, we detail the simulation environments and
the algorithm. 

\paragraph{Environment Setup}
\label{sec:setup}
We consider a 2D rectangular
environment 
discretized by $0.1 \times 0.1$ unit square grid cells (see
Figure~\ref{fig:extra_sim_environment} for a map layouts).
The ground truth map $\map^{*}$ consists solely of free-space
($0$ occupancy value) and obstacles ($100$ occupancy value). This map was used
in the DARPA Subterreanean challenge~\cite{JGR:20}. In
Figure~\ref{fig:extra_sim_environment}, the blue/dark areas ($0$) represent
free space, the yellow areas ($100$) represent known obstacles and other values indicate uncertain areas. We construct
the initial occupancy map $\map$ by adding noise to each cell value of the ground truth DARPA subT map $\map^*$.
The environment is divided into four quadrants which contain different
levels of noise, sampled from a uniform distribution with different
intervals from $[0,50]$ (top-left quadrant), to $[0,80]$ (top-right
quadrant), and others with $[0,30]$ (bottom-right) and $[0,20]$
(bottom-left) (see Figure~\ref{fig:extra_sim_environment} for reference). This is to simulate the fact that different sections of the DARPA subT map $\map^*$ can have different initial uncertainties. We
do not add noise to the extreme areas of the map so that the robot
remains within map boundaries.
The initial robot
positions are sampled uniformly from the free space in
$\map^*$. One such sample initial condition is shown in Figure~\ref{fig:extra_sim_environment}, where the robot positions $x_i(0)$ is represented by dots beside the robot ID ($i$).
In order to simulate
heterogeneity among robots we assign their $\alpha$ values from the
following density function
\begin{align*}
	\mathrm{PWC}[a,b] \ldef 
		\begin{cases}
			\frac{1}{b-a} \one_{[a,b]}, & \mathrm{if} \,\, 1 \notin (a,b)\,; \\
			\frac{1}{2(1-a)} \one_{[a,1]} + \frac{1}{2(b-1)} \one_{[1,b]}, & \mathrm{if} \,\, 1 \in (a,b)\,;
		\end{cases}
\end{align*} 
where $\one_{\set}$ is the indicator function for the set
$\set$. We use different ranges of $(a,b)$ as simulation parameters to check mapping performance. For instance, if $1<a<b$ (resp. $a<b<1$), the all robots are uncertainty insensitive (resp. averse); and if $a<1<b$, we have mixed behavior.
To that extent, in Figure~\ref{fig:extra_sim_environment}, we color robot $i \in \agt$ red if $\alpha_i > 1$ and blue if $\alpha_i < 1$.
We use 10 robots randomly
selected from $\mathrm{PWC}[a,b]$ to perform simulations. 

\begin{algorithm}[t]
\SetAlgoLined
Input : $\map$, $x_i$, $K$ ; Output : $\focc(\grid)$\\
$\texttt{taskBuffer}_i \gets \varnothing$ 
\tcp*{\scriptsize frontiers to visit}
$\tsk_i \gets \texttt{getFrontier}(\mathcal{M},K)$
\tcp*{\scriptsize \hspace{-2.5ex} initial frontiers}
\While{$\tsk_i \neq \varnothing$}{
	\If{$\texttt{taskBuffer}_i = \varnothing$}{
		$\{w_i^q\}_{q \in \tsk} \gets \texttt{communicate}(\tsk_i, \neigh_i)$\\
		$\texttt{tour} \gets \texttt{pick}(\{q \,|\, w_i^q =1\}, N)$\\
		$\texttt{tour} \gets \texttt{travelingSalesman}(\texttt{tour})$\\
		$\texttt{taskBuffer}_i \gets \texttt{tour}$ \\
	}
	$\texttt{path} \gets \texttt{RRT}(x_i,\texttt{taskBuffer}_i[0])$ \\
	Move to $\texttt{taskBuffer}_i[0]$ following $\texttt{path}$ and map along the way\\
	$x_i \gets \texttt{taskBuffer}_i[0]$\\
	Remove $\texttt{taskBuffer}_i[0]$\\
	$\tsk_i \gets \texttt{getFrontier}(\mathcal{M},K)$
}
\caption{Behavioral explorer}
\label{alg:explorer}
\end{algorithm}

\paragraph{Exploration pipeline}
\label{sec:pipeline}
The formal description of the exploration pipeline is given in
Algorithm~\ref{alg:explorer}. Each robot $i \in \agt$ runs
Algorithm~\ref{alg:explorer} separately and addresses the
communication call among its neighbors $\neigh_i$ whenever needed. The
neighbors of a robot at any time is the set of robots that have common
frontiers.  Each robot initializes an empty task buffer to keep track
of the frontiers they visit.
 
First, robot $i \in \agt$ updates a circular area
$\mathcal{Q} = \mathbf{B}^{x_i}_r$ in $\grid$, with a radius $r$ and
centered around its current position $x_i$. 
We use three
different mapping radii $r \in \{2,3,4\}$ to capture different sensor
ranges. Three levels of mapping noise $\sigma^m \in \{0,1,2\}$ are
employed with the following meaning: \textit{a)} In perfect mapping
($\sigma^m = 0$) each cell in $\mathcal{Q}$ is updated to the ground
truth.  \textit{b)} In imperfect mapping ($\sigma^m = 1$) the
occupancy values in $\mathcal{Q}$ are reduced (resp. increased)
randomly with a random number generated by a uniform distribution with
range $[0,35]$ for each cell in $\mathcal{Q}$ that is a free-space
(resp. obstacle).  \textit{c)} In highly imperfect mapping
($\sigma^m = 2$), the occupancy values are updated similarly with
random numbers generated by a uniform distribution with range
$[0,15]$.

Second, frontiers $\tsk_i$ are obtained as those cells in robot $i$'s
map $\map_i$ with occupancy values $<2$ and a non-negative gradient
w.r.t.~to the cells around them; see \cite{AS-CN-SM:24-ral}. Each
robot only considers frontiers in a certain radius around it. We set
this frontier radius initially to $10$ times the sensing radius; see
Figure~\ref{fig:extra_sim_environment}. The robot can increase the frontier
radius if it is not able to see any frontier within this radius. This
helps in reducing compute time and also communication cost.

If the robot has an empty task buffer, it communicates with its
neighbors to fill up this task buffer. For each $q \in \mathcal{F}_i$,
we calculate the information gain by considering
$\mathcal{Q}=\mathbf{B}^{q}_r$. That is, for each cell in
$\mathcal{Q}$, the information gain per cell is given by the
BE in these cells and the total information gain
$\nu_i$ is the sum of these values. Then, for each
$q \in \tsk_i$, the utility is calculated
using~\eqref{eq:beh_utility}. 
Using this utility, the robots run rounds of the d-PBRAG algorithm,
sufficiently long, to allocate the available frontiers on their map
among themselves. At the end of the communication rounds, robot $i$
considers the frontiers $q \in \tsk_i$ that have weights $w^q_i = 1$
and chooses at most $N$ (we set $N=14$) frontiers. The
robot then finds a traveling salesman tour (TST) among these $N$
frontiers.

Finally, the robot moves from $x_i$ to the first frontier $q_0$ in the
TST using a path computed using the RRT planner. The robot also maps
(according to first step) all possible circular regions within its
sensor range along the path. The robot then removes $q_0$ from the TST
and continues the algorithm.

\begin{algorithm}[t]
\SetAlgoLined
Input : $\tsk_i$, $\neigh_i$ ;
Output : $\{w_i^q\}_{q \in \tsk}$ \\
$w_i^q \gets 1$, $\forall q \in \tsk$
\tcp*{\scriptsize initialize all weights to 1}
$z_i^q \gets \rf(q) \phi(q)$, $\forall q \in \tsk$
\tcp*{\scriptsize utility using~\eqref{eq:beh_utility}}
Run~\eqref{eq:dist_pbrag} sufficiently long using $\tsk_i$, $\neigh_i$
\tcp*{\scriptsize d-PBRAG}
\caption{\texttt{communicate}}
\end{algorithm}

\paragraph{Results and Discussion}
\label{sec:results}

\begin{figure*}
	\def \fig_scale{0.35}
	\begin{center}
	\begin{minipage}{0.3\textwidth}
		\subcaptionbox{$r = $2,  $\sigma^m = $ 2\label{fig:efficiency_2_noise_2}}{\includegraphics[trim=27pt 0pt 0pt 19pt, clip, width=\textwidth]{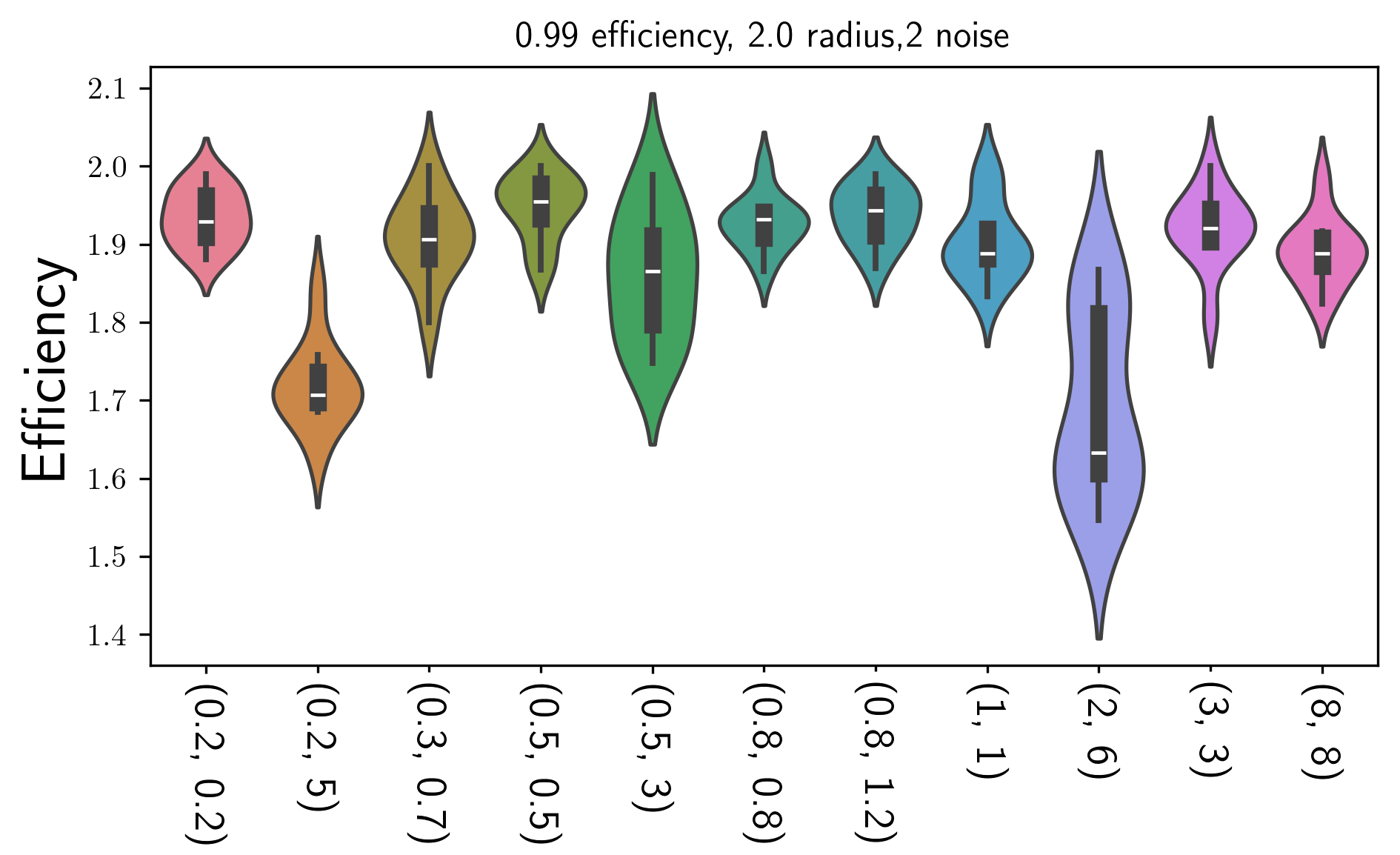}}
	\end{minipage}
	\begin{minipage}{0.3\textwidth}
		\subcaptionbox{$r = $3,  $\sigma^m = $ 2\label{fig:efficiency_3_noise_2}}{\includegraphics[trim=27pt 0pt 0pt 20pt, clip, width=\textwidth]{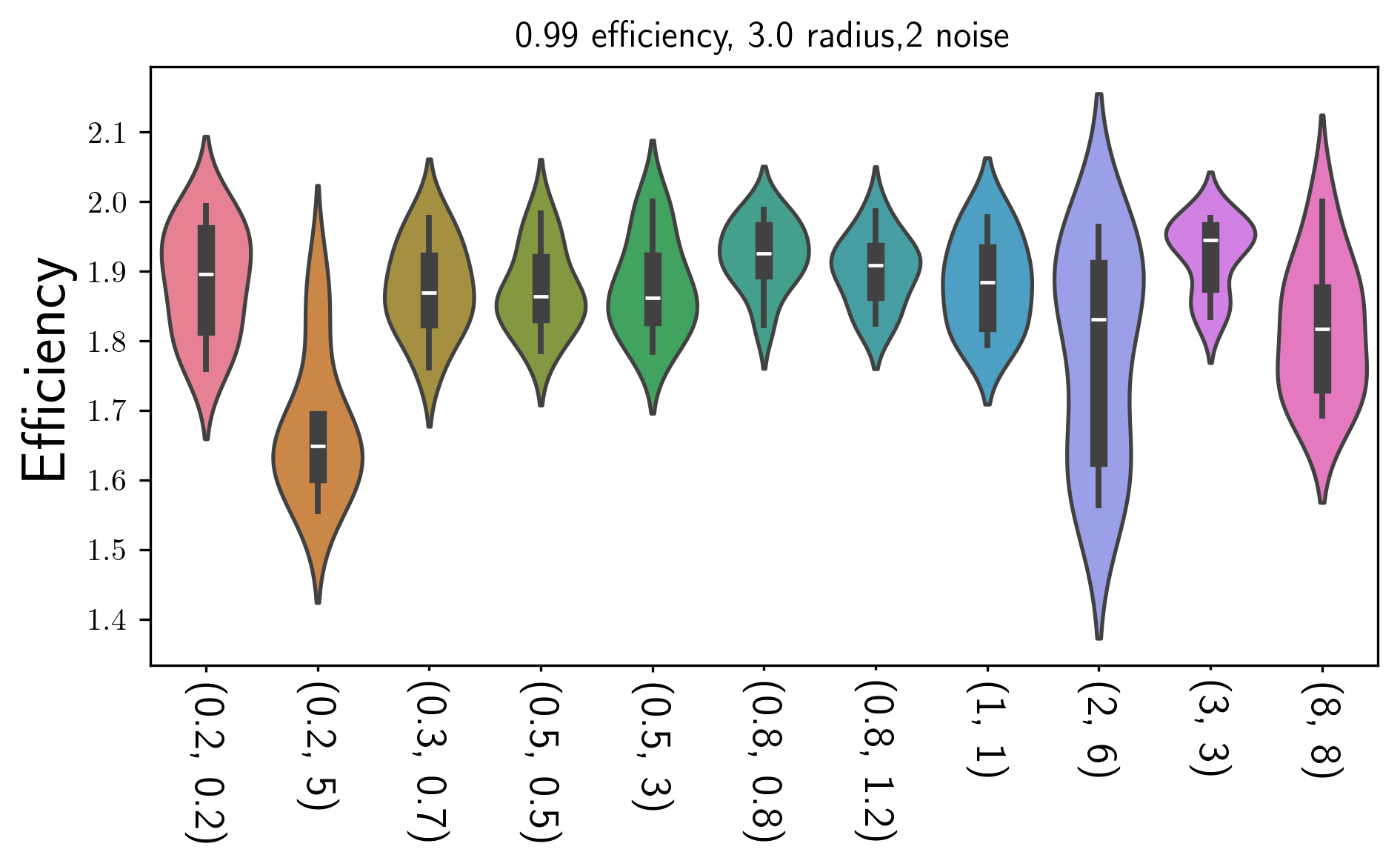}}
	\end{minipage}
	\begin{minipage}{0.3\textwidth}
		\subcaptionbox{$r = $4,  $\sigma^m = $ 2\label{fig:efficiency_4_noise_2}}{\includegraphics[trim=27pt 0pt 0pt 20pt, clip, width=\textwidth]{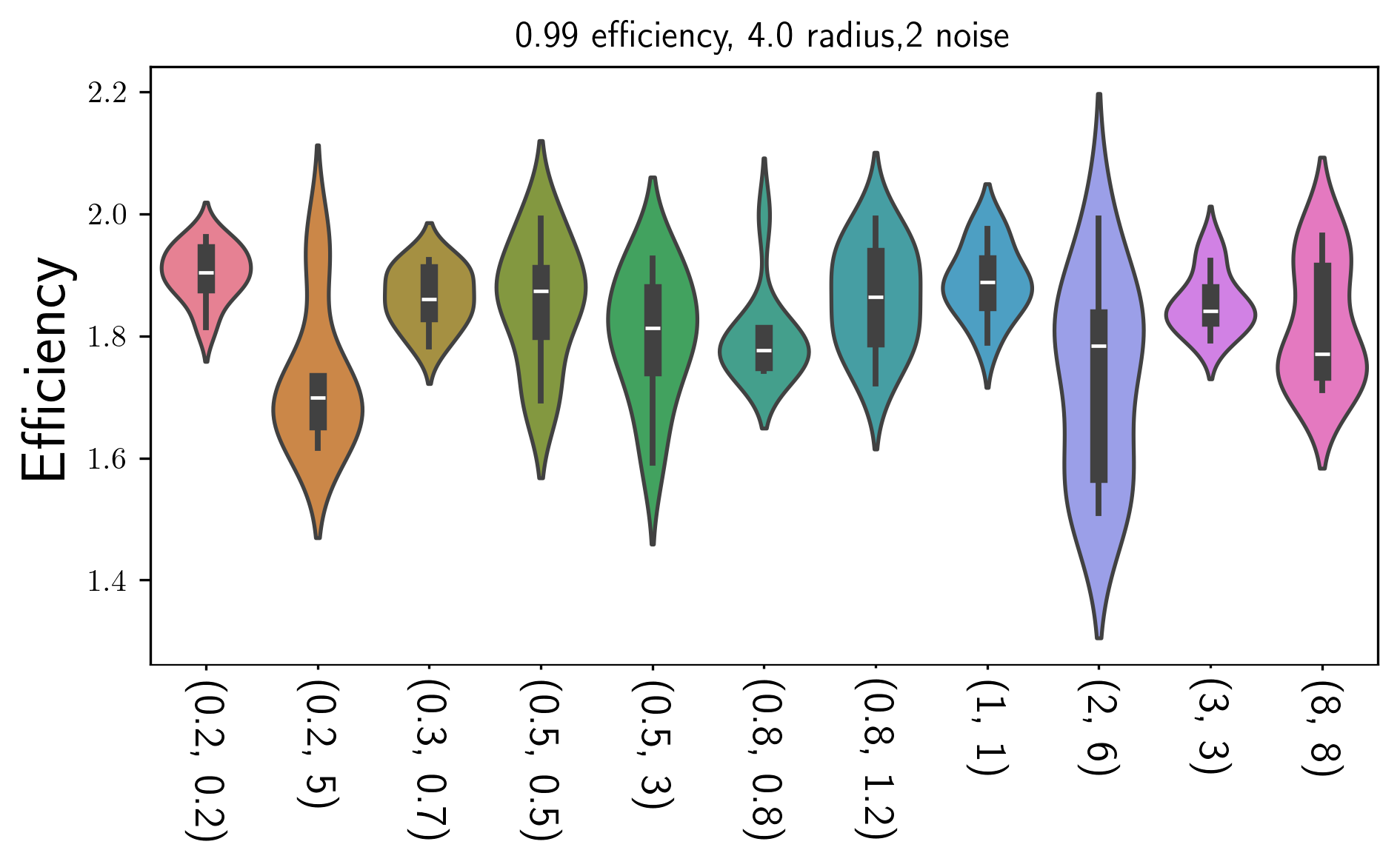}}
	\end{minipage}\\
	
	\begin{minipage}{0.3\textwidth}
		\subcaptionbox{$r = $2,  $\sigma^m = $ 1\label{fig:efficiency_2_noise_1}}{\includegraphics[trim=27pt 0pt 0pt 20pt, clip, width=\textwidth]{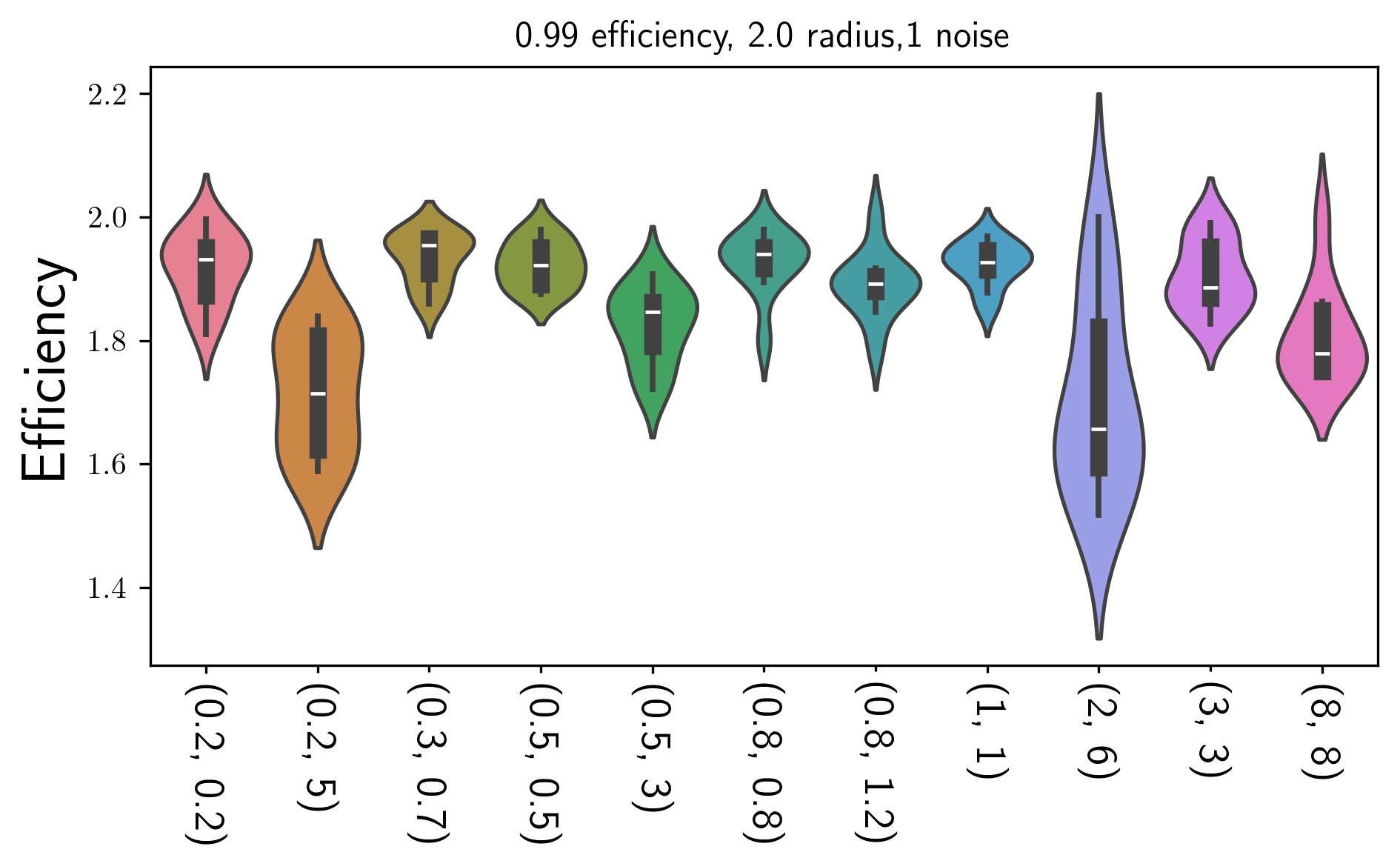}}
	\end{minipage}
	\begin{minipage}{0.3\textwidth}
		\subcaptionbox{$r = $3,  $\sigma^m = $ 1\label{fig:efficiency_3_noise_1}}{\includegraphics[trim=27pt 0pt 0pt 20pt, clip, width=\textwidth]{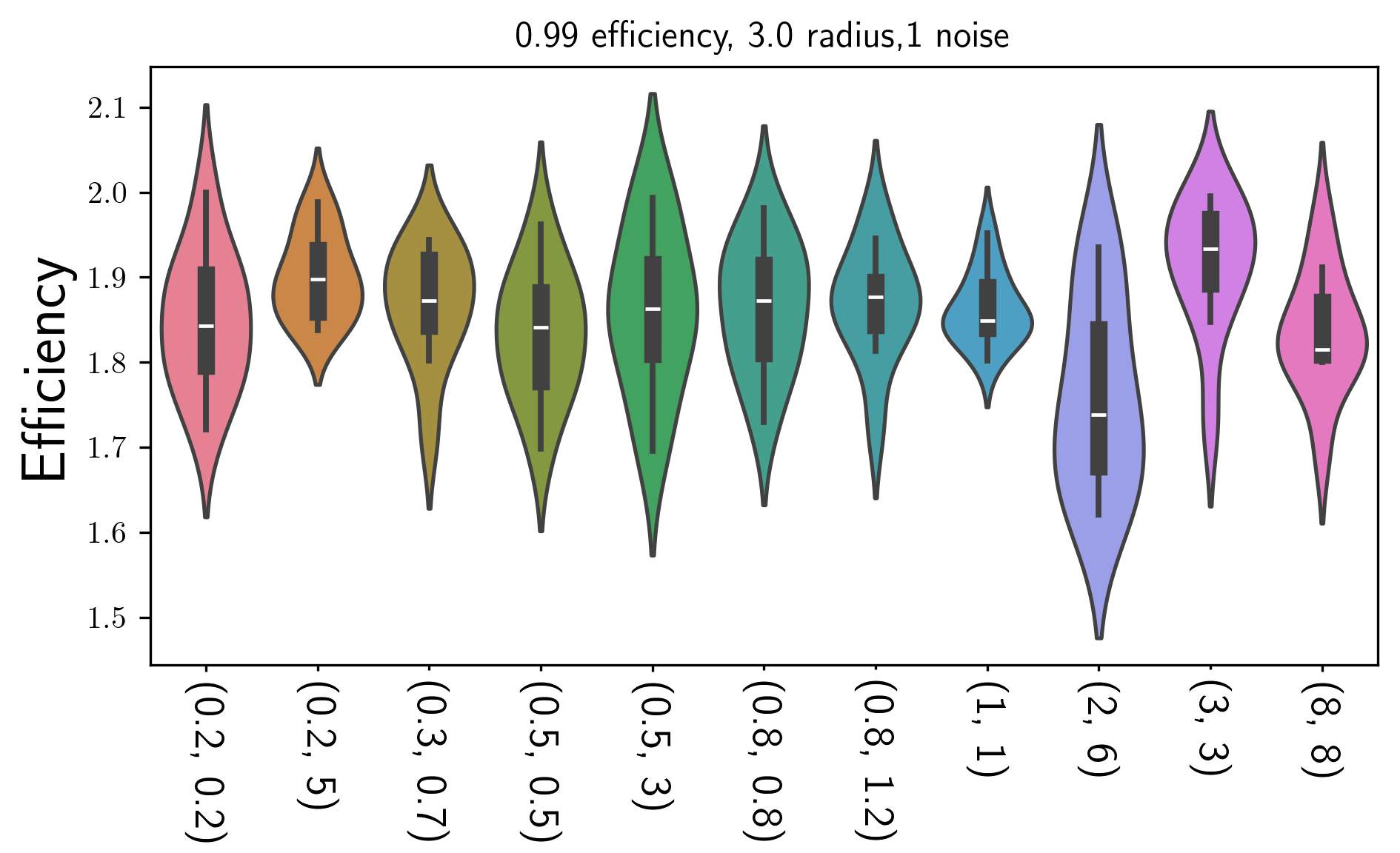}}
	\end{minipage}
	\begin{minipage}{0.3\textwidth}
		\subcaptionbox{$r = $4,  $\sigma^m = $ 1\label{fig:efficiency_4_noise_1}}{\includegraphics[trim=27pt 0pt 0pt 20pt, clip, width=\textwidth]{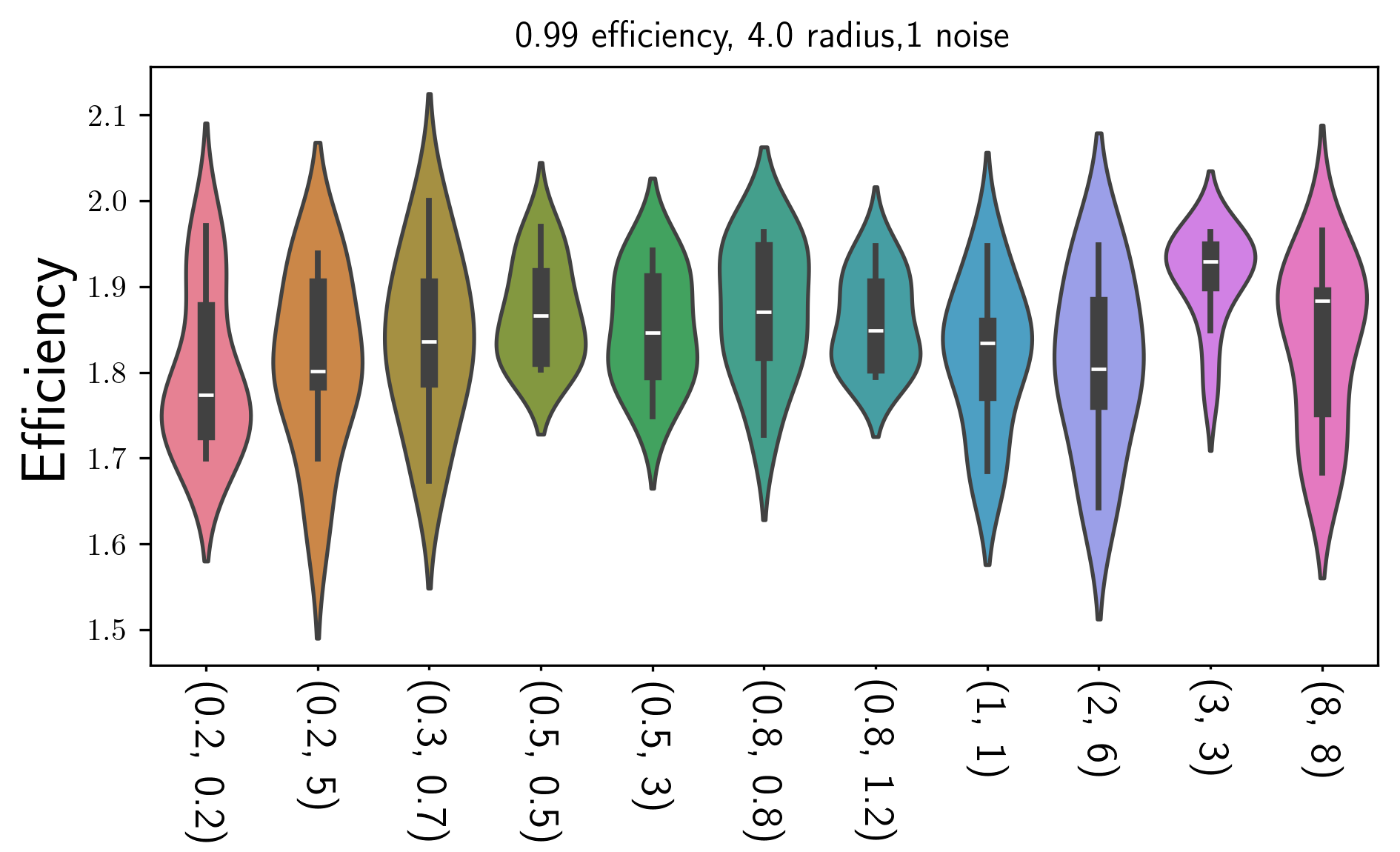}}
	\end{minipage} \\
	
	\begin{minipage}{0.3\textwidth}
		\subcaptionbox{$r = $2,  $\sigma^m = $ 0\label{fig:efficiency_2_noise_0}}{\includegraphics[trim=27pt 0pt 0pt 19pt, clip, width=\textwidth]{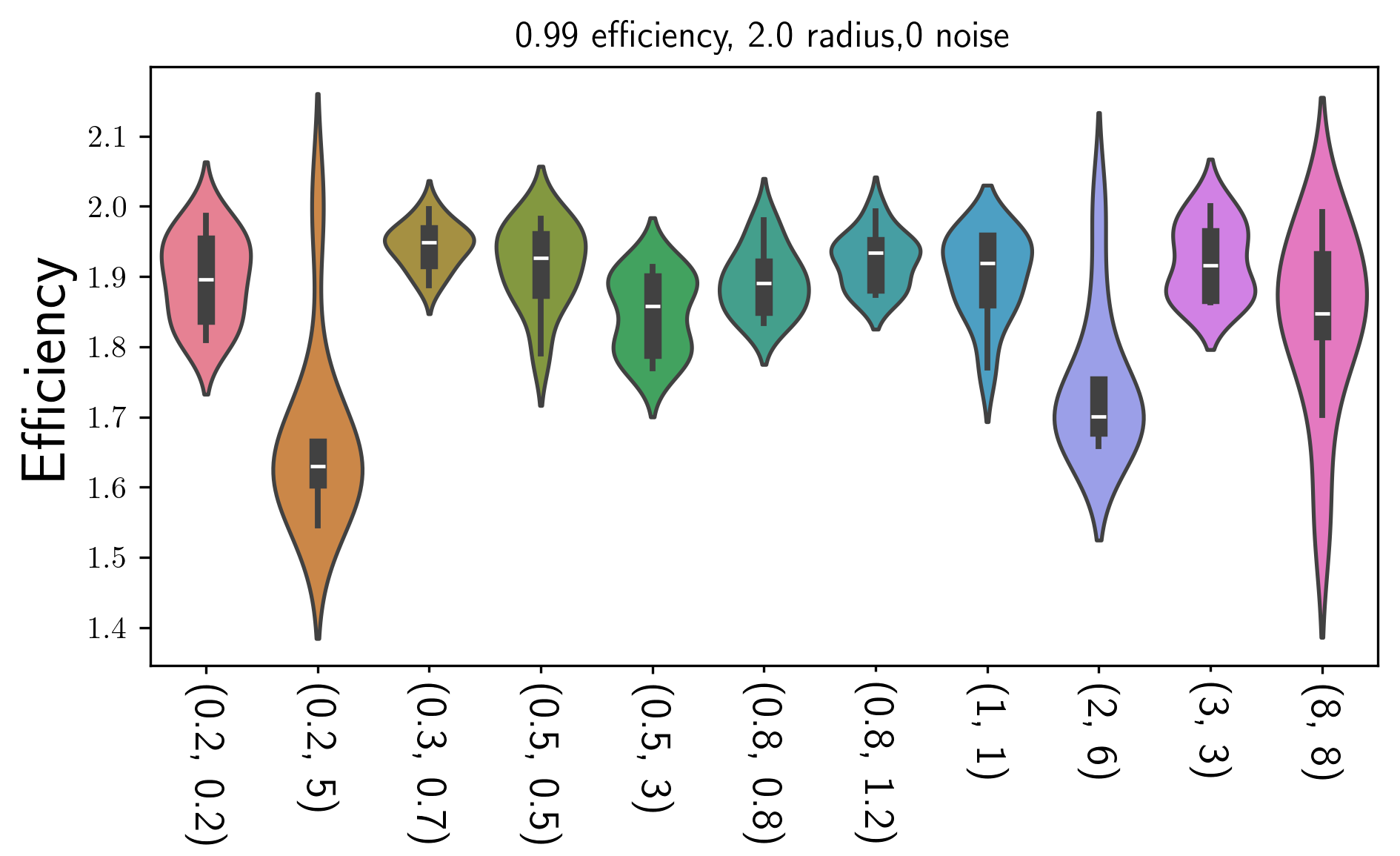}}
	\end{minipage}
	\begin{minipage}{0.3\textwidth}
		\subcaptionbox{$r = $3,  $\sigma^m = $ 0\label{fig:efficiency_3_noise_0}}{\includegraphics[trim=27pt 0pt 0pt 20pt, clip, width=\textwidth]{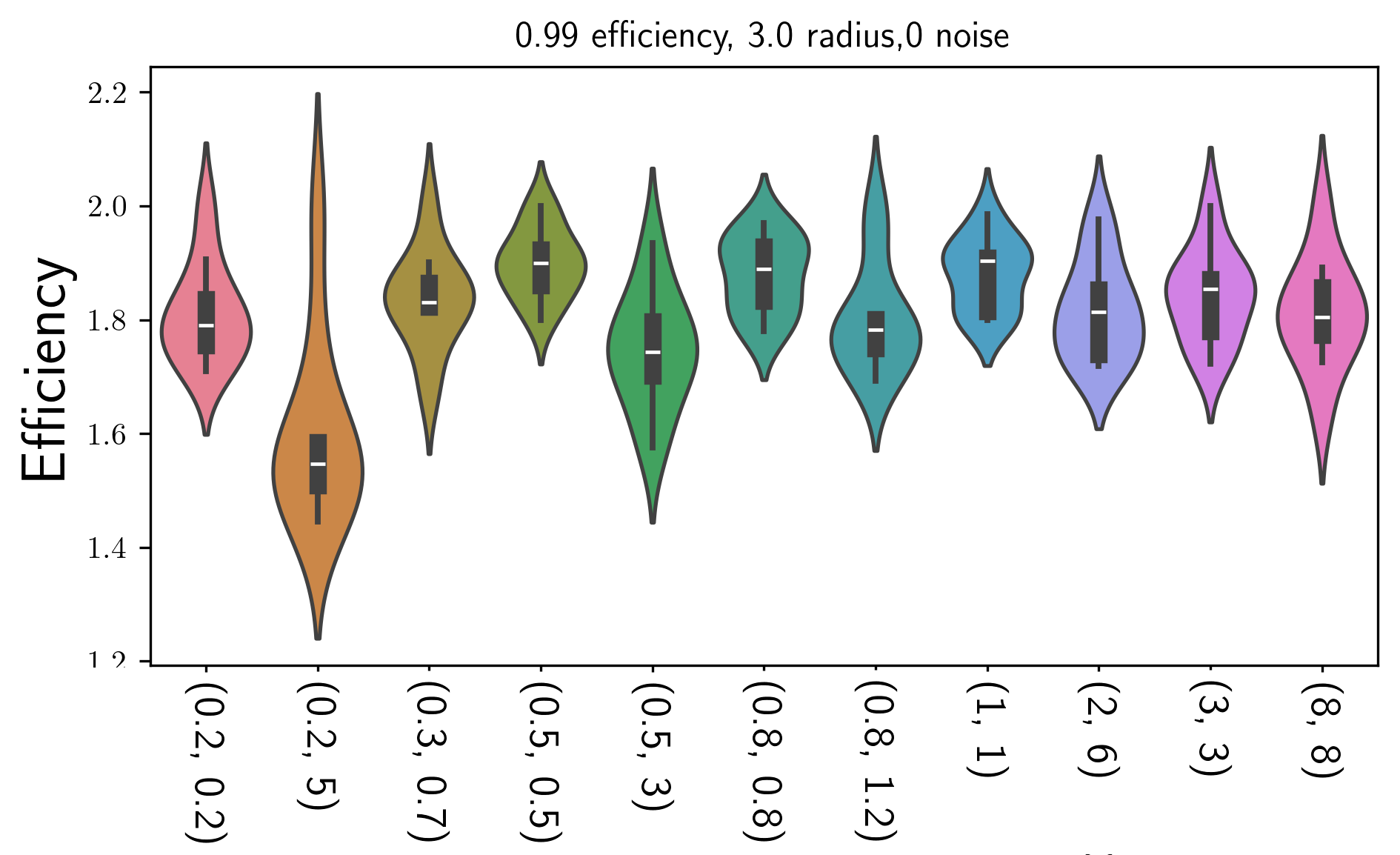}}
	\end{minipage}
	\begin{minipage}{0.3\textwidth}
		\subcaptionbox{$r = $4,  $\sigma^m = $ 0\label{fig:efficiency_4_noise_0}}{\includegraphics[trim=27pt 0pt 0pt 20pt, clip, width=\textwidth]{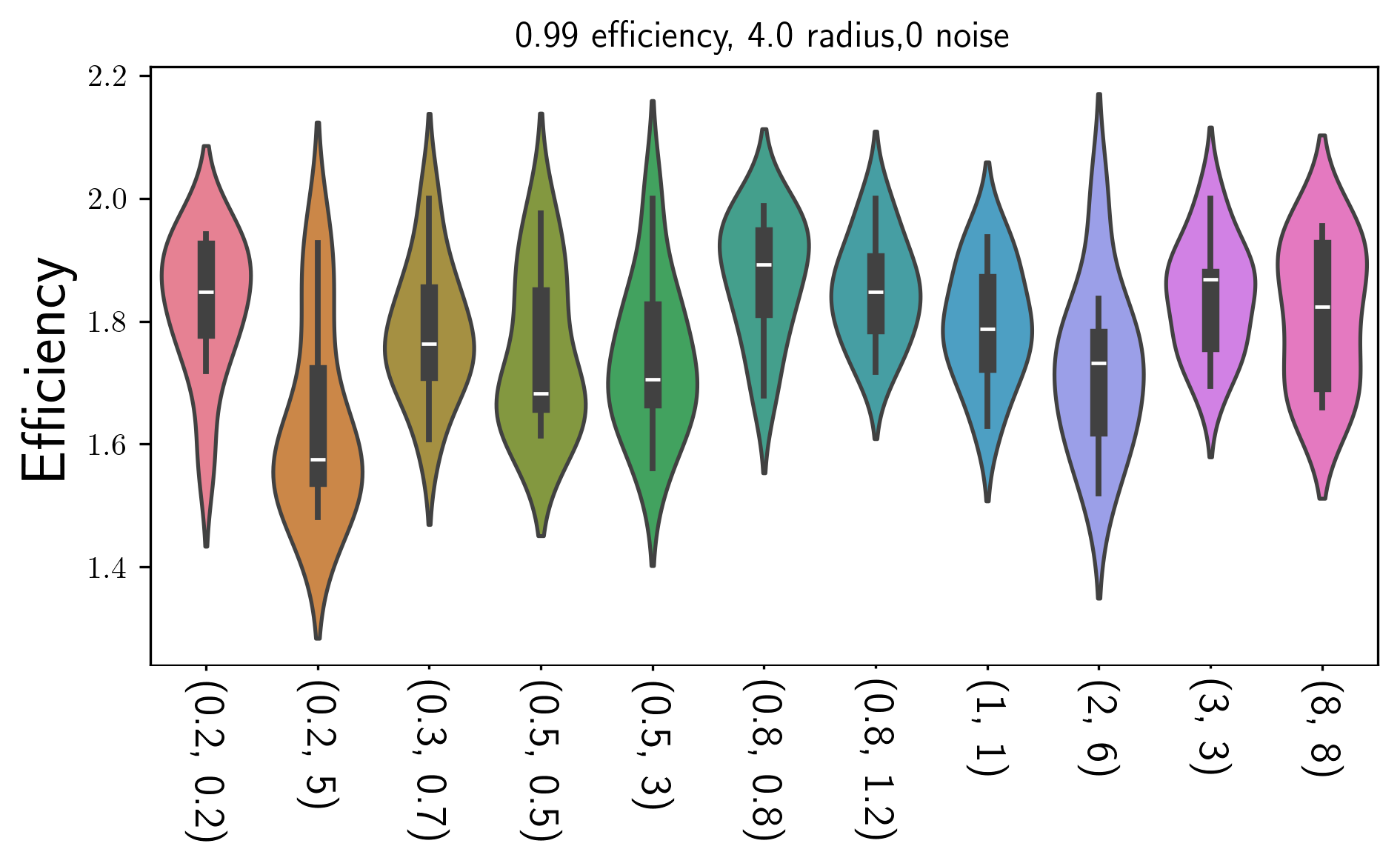}}
	\end{minipage}
	\end{center}
	\caption{Cost for DARPA subT map. The rows from top to bottom have $\sigma^m = 2,1,0$ resp. The columns from left to right have $r = 2,3,4$ resp.}
	\label{fig:darpa_subT}
\end{figure*}

\begin{figure*}
	\def \fig_scale{0.35}
	\begin{center}
	\begin{minipage}{0.3\textwidth}
		\subcaptionbox{$r = $2.0, $\sigma^m = 2$}{\includegraphics[trim=23pt 0pt 0pt 25pt, clip, width=\textwidth]{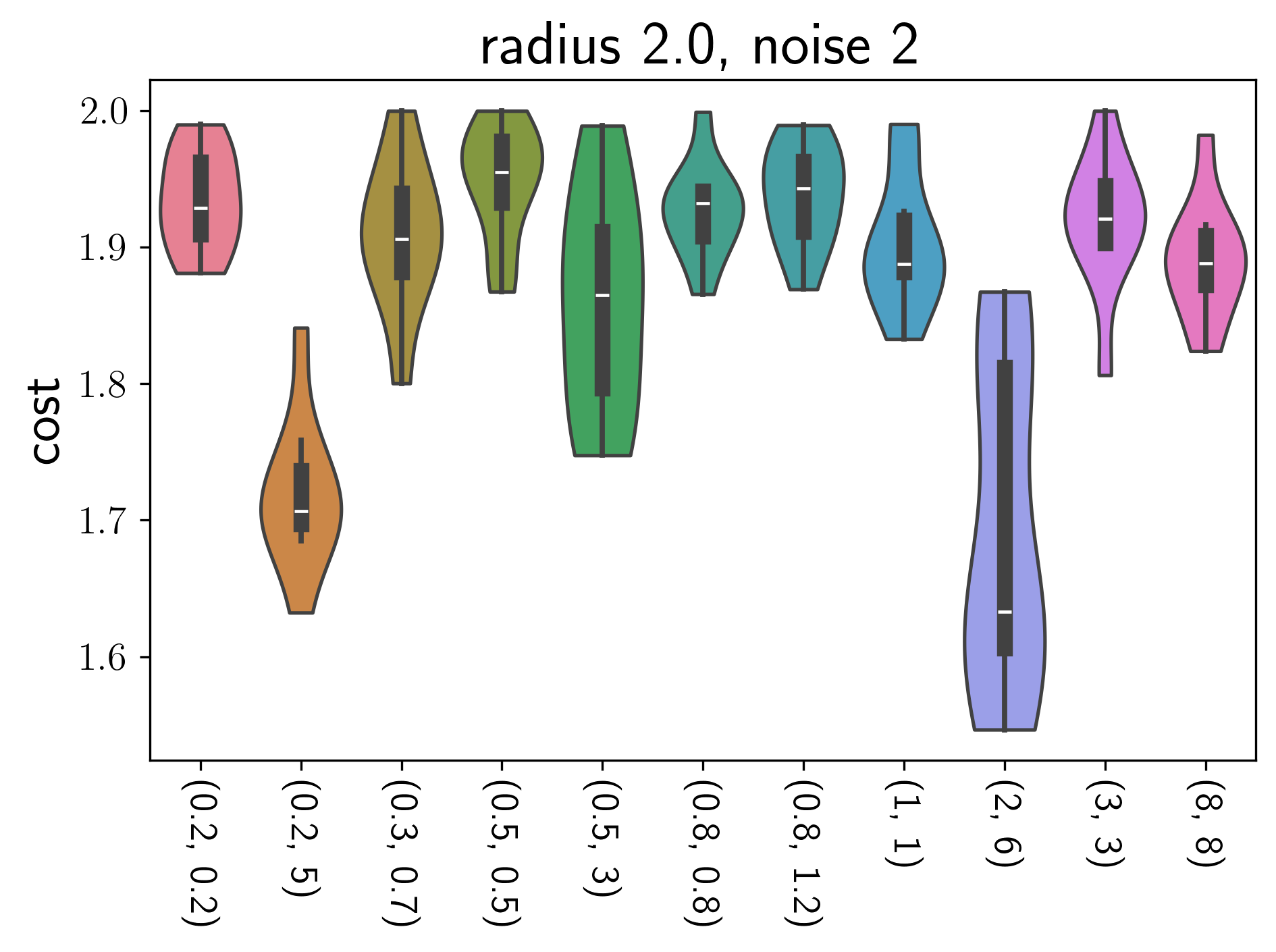}}
	\end{minipage}
	\begin{minipage}{0.3\textwidth}
		\subcaptionbox{ $r = $3.0, $\sigma^m = 2$}{\includegraphics[trim=23pt 0pt 0pt 25pt, clip, width=\textwidth]{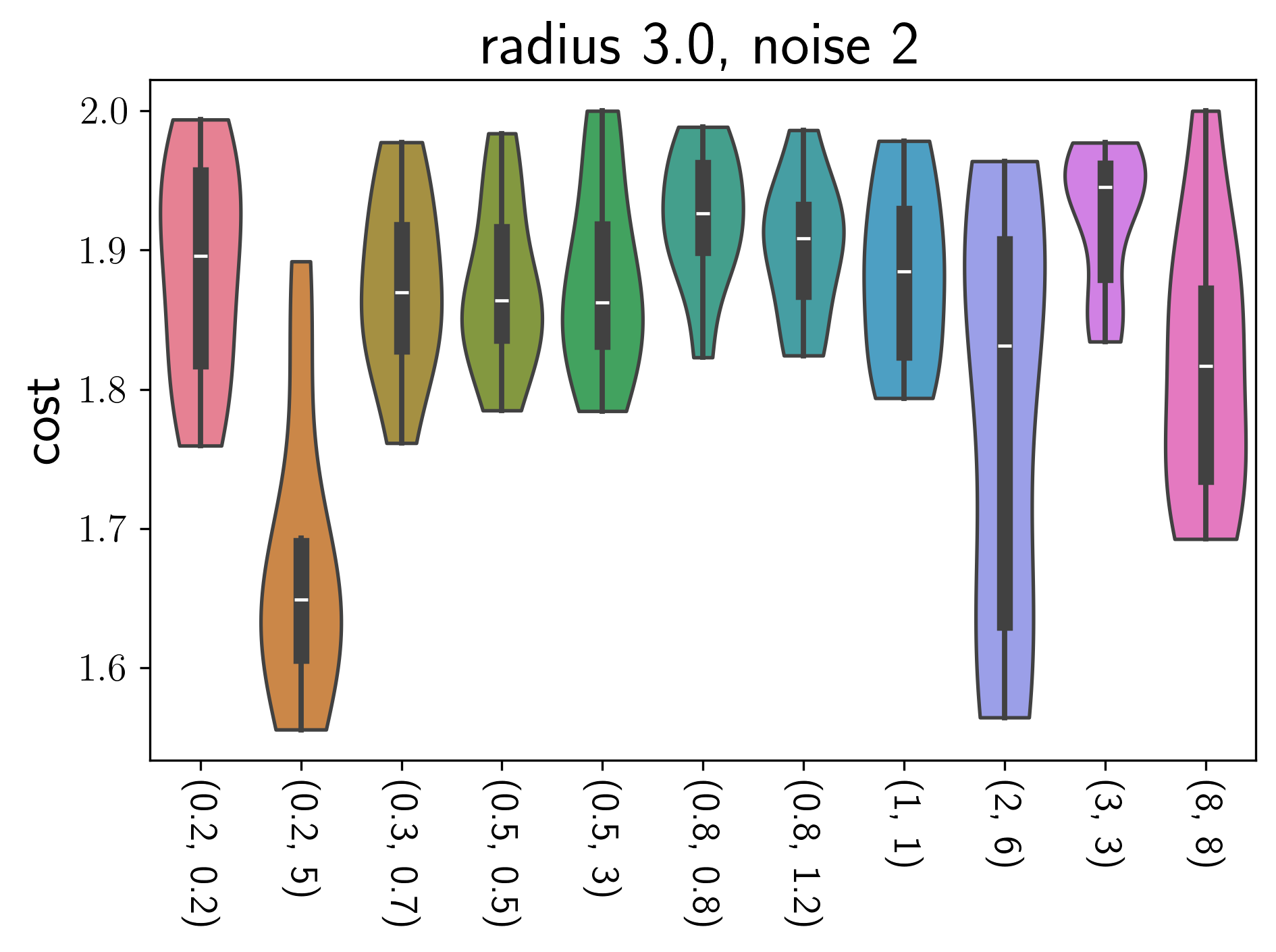}}
	\end{minipage}
	\begin{minipage}{0.3\textwidth}
		\subcaptionbox{ $r = $4.0, $\sigma^m = 2$}{\includegraphics[trim=23pt 0pt 0pt 25pt, clip, width=\textwidth]{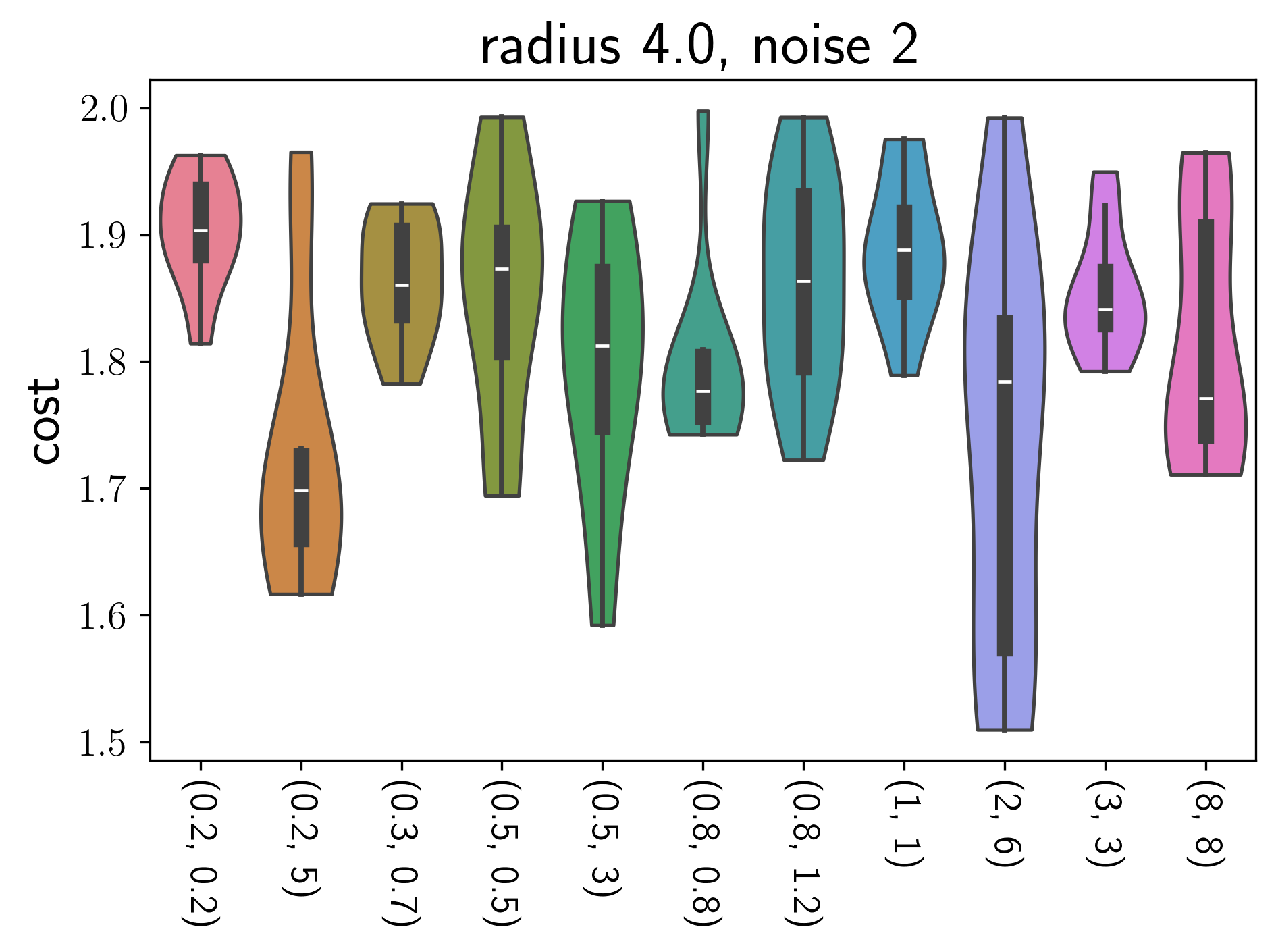}}
	\end{minipage} \\
	
	\begin{minipage}{0.3\textwidth}
		\subcaptionbox{ $r = $2.0, $\sigma^m = 1$}{\includegraphics[trim=23pt 0pt 0pt 25pt, clip, width=\textwidth]{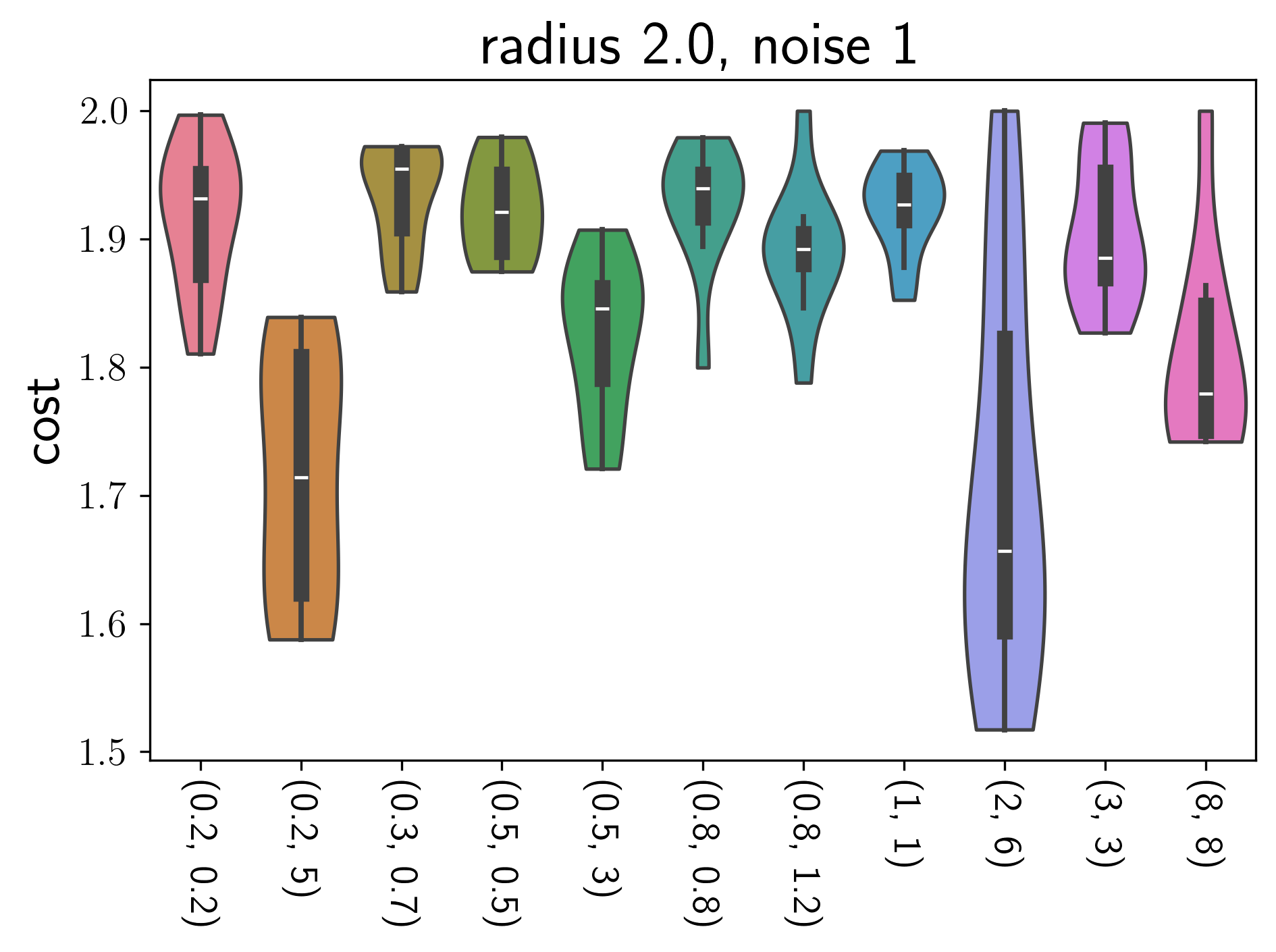}}
	\end{minipage}
	\begin{minipage}{0.3\textwidth}
		\subcaptionbox{ $r = $3.0, $\sigma^m = 1$}{\includegraphics[trim=23pt 0pt 0pt 25pt, clip, width=\textwidth]{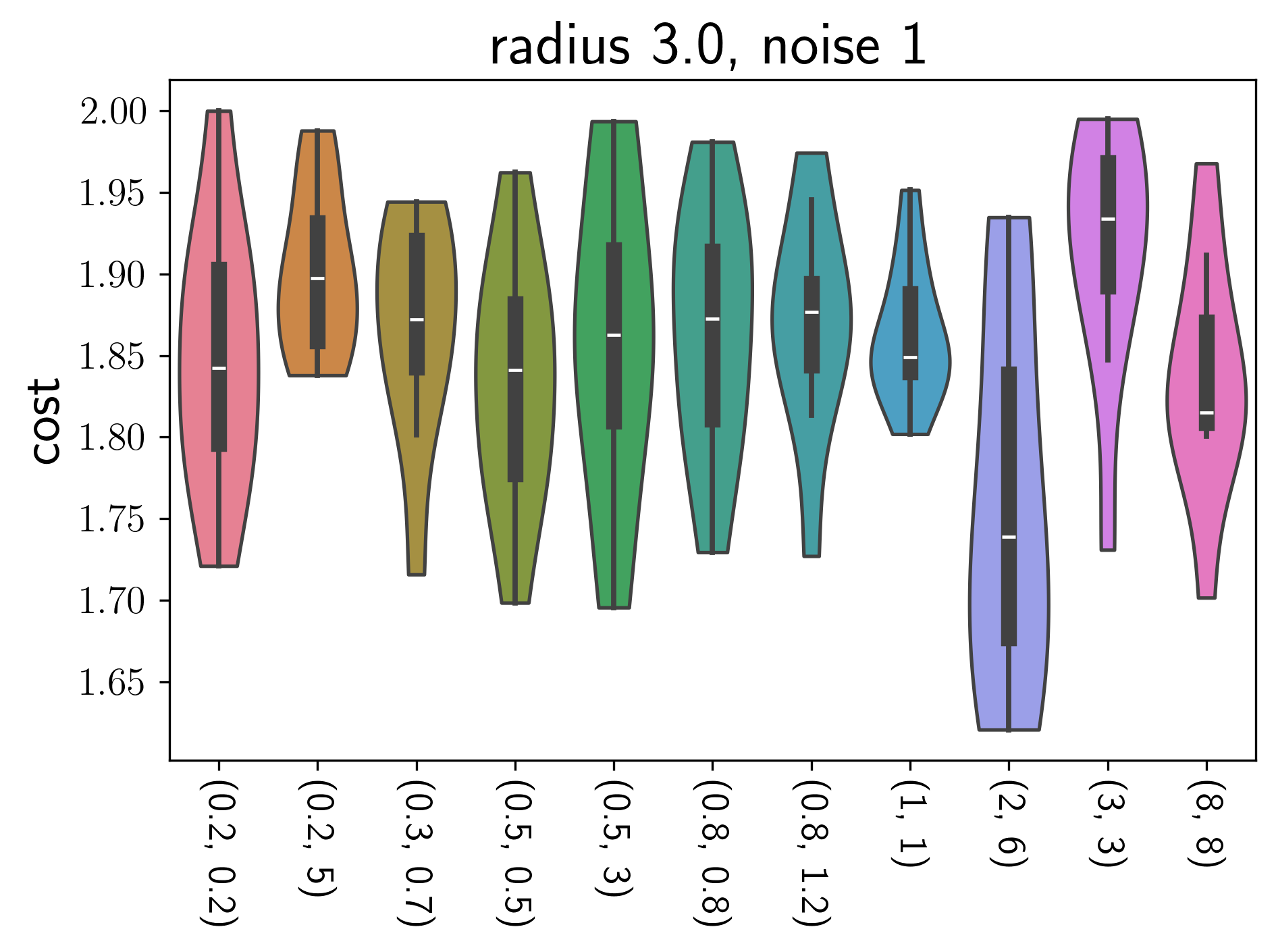}}
	\end{minipage}
	\begin{minipage}{0.3\textwidth}
		\subcaptionbox{ $r = $4.0, $\sigma^m = 1$}{\includegraphics[trim=23pt 0pt 0pt 25pt, clip, width=\textwidth]{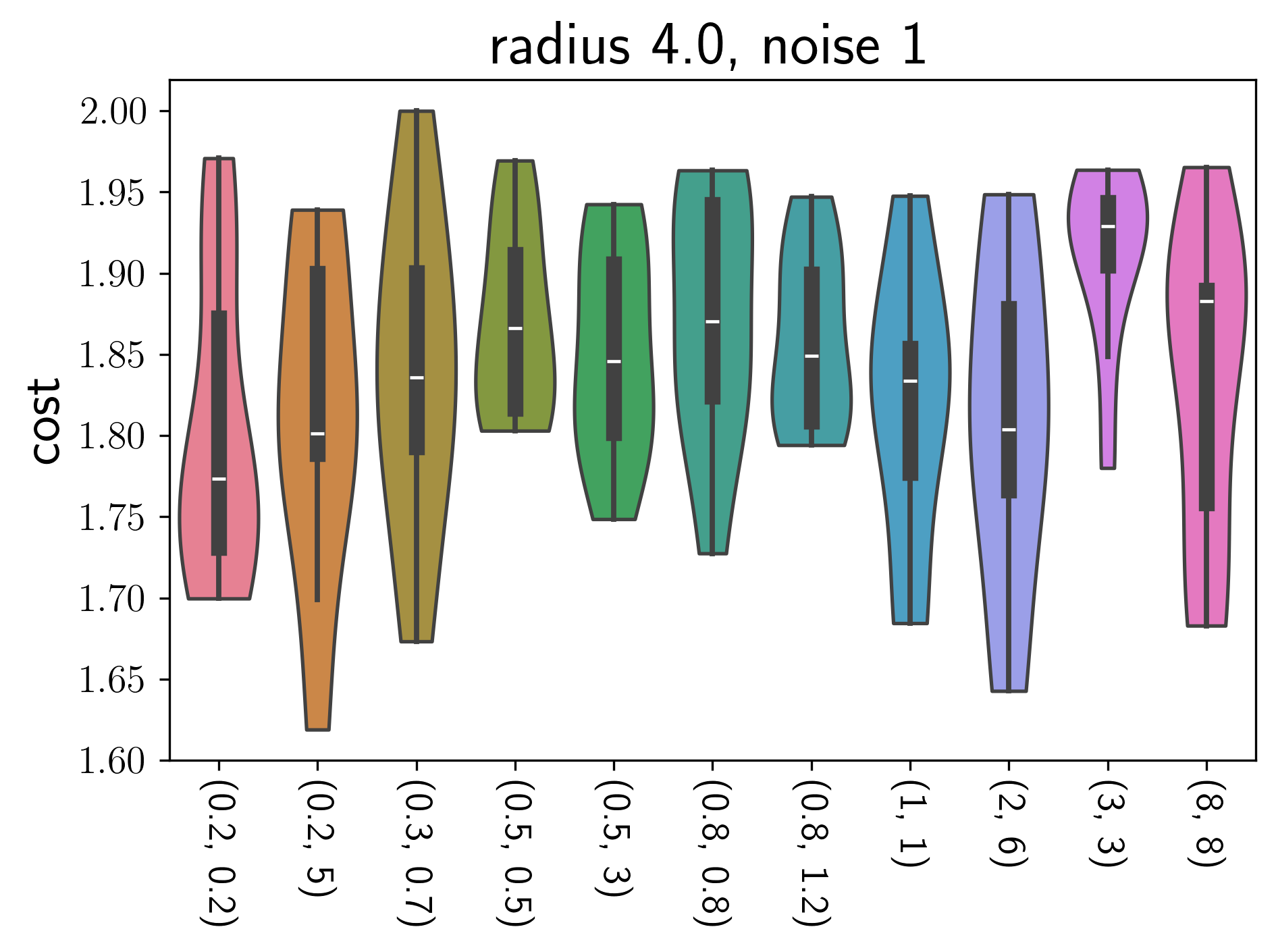}}
	\end{minipage} \\
	
	\begin{minipage}{0.3\textwidth}
		\subcaptionbox{ $r = $2.0, $\sigma^m = 0$}{\includegraphics[trim=23pt 0pt 0pt 25pt, clip, width=\textwidth]{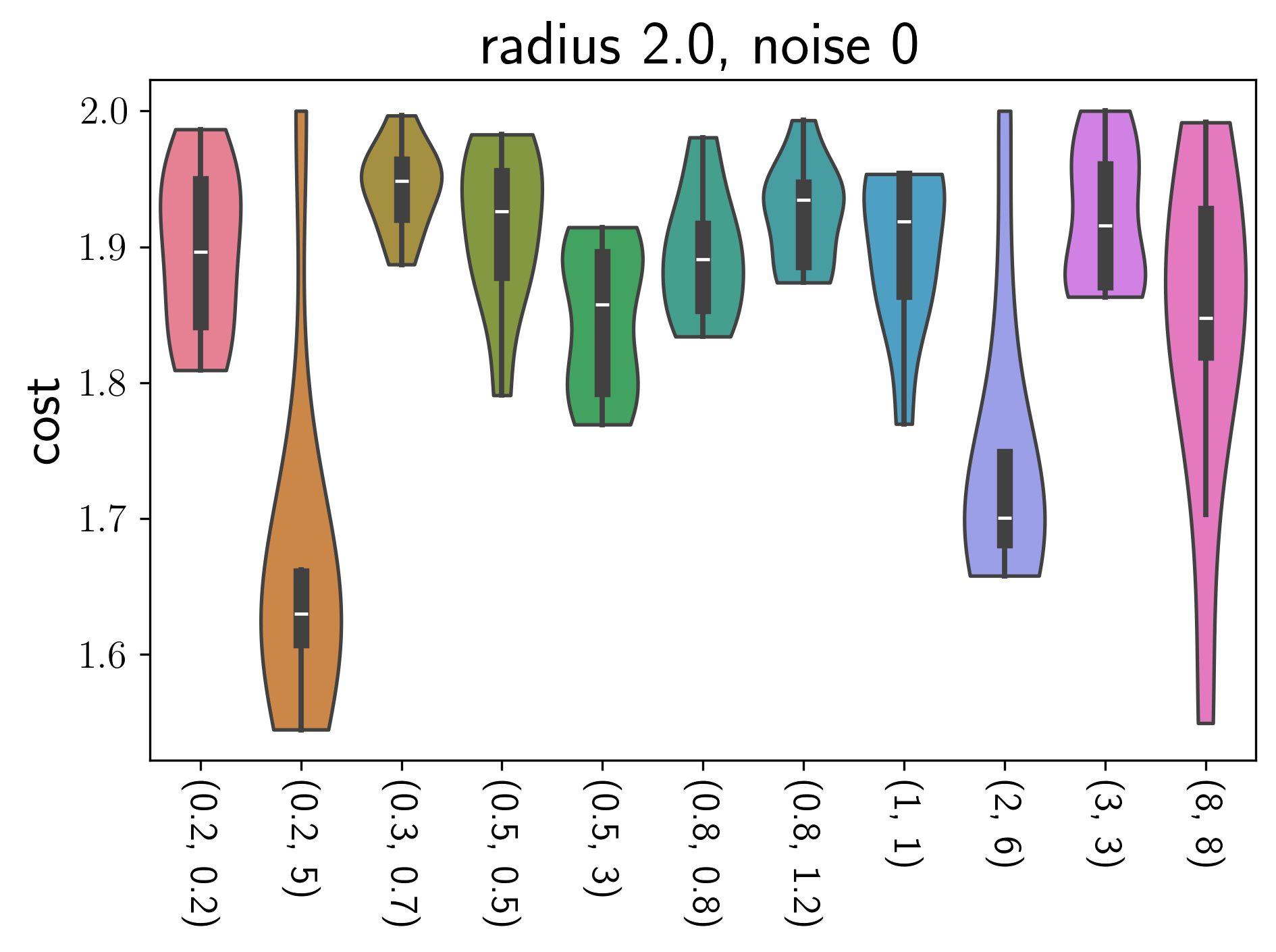}}
	\end{minipage}
	\begin{minipage}{0.3\textwidth}
		\subcaptionbox{ $r = $3.0, $\sigma^m = 0$}{\includegraphics[trim=23pt 0pt 0pt 25pt, clip, width=\textwidth]{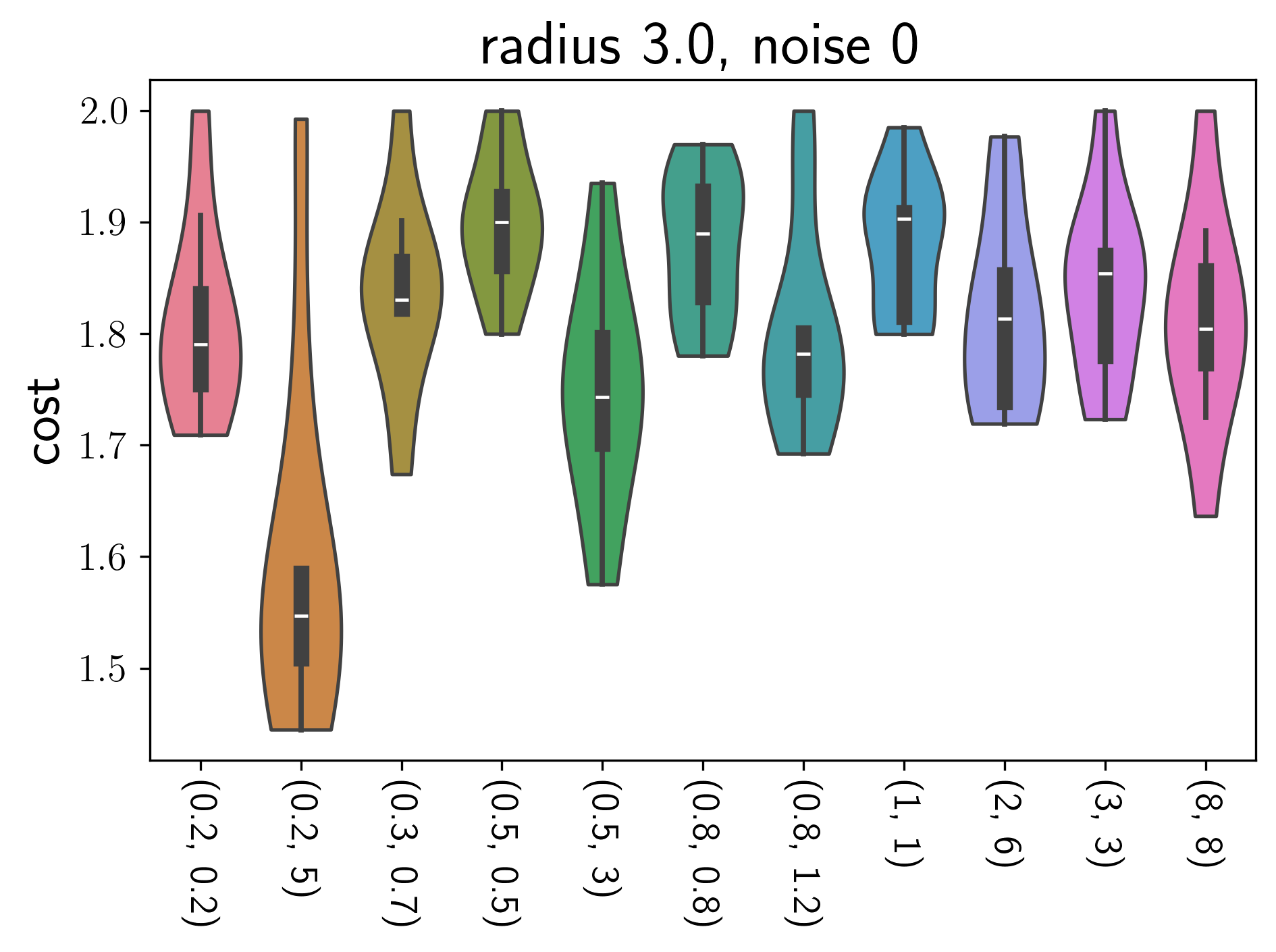}}
	\end{minipage}
	\begin{minipage}{0.3\textwidth}
		\subcaptionbox{ $r = $4.0, $\sigma^m = 0$}{\includegraphics[trim=23pt 0pt 0pt 25pt, clip, width=\textwidth]{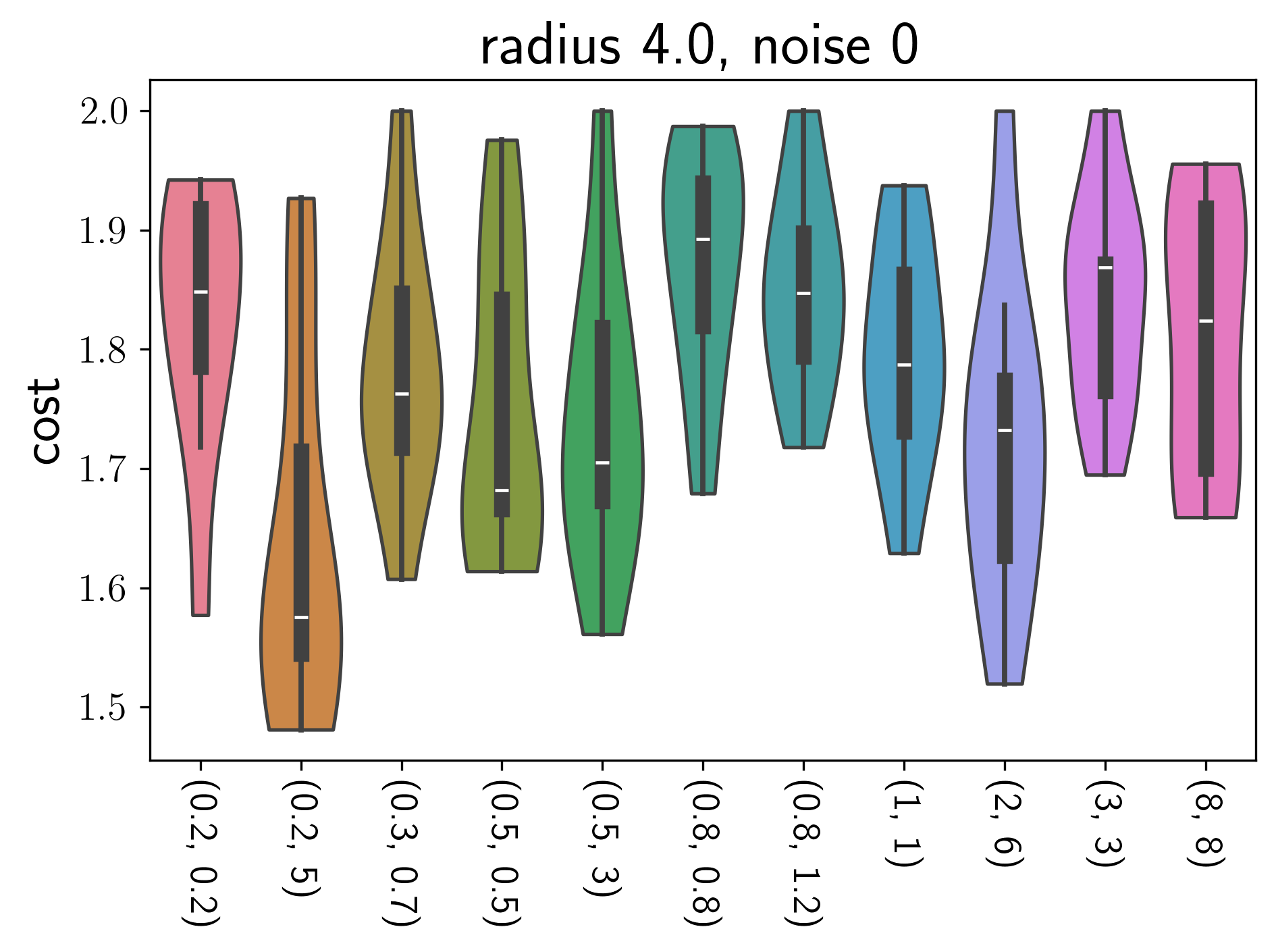}}
	\end{minipage}
	\end{center}
	\caption{Cost for urban circuit map. The rows from top to bottom have $\sigma^m = 2,1,0$ resp. The columns from left to right have $r = 2,3,4$ resp.}
	\label{fig:uc_32r}
\end{figure*}

We simulate 11 different $\alpha$ ranges to capture a wide range of
behaviors. For each $\alpha$ range, each sensing radius $r \in \{2,3,4\}$ and noise levels $\sigma^m \in \{0,1,2\}$ we perform 10 simulations. We used two different maps for these simulations (see Figure~\ref{fig:extra_sim_environment}). This brings the total number of simulations to 1980.  The results in
Figure~\ref{fig:entropy_and_path_length} show the distribution of the
number of iterations taken to explore $99 \%$ of the map.  The Y-axis
indicates the distribution of number of iterations used with a
particular group to reduce at least $99 \%$ of initial entropy $\mathcal{H}(0)$. 
Wider (resp. narrower) violin denotes that more (resp. fewer) trials gave similar number of iterations. 

First, we report the trade-off between time taken for entropy
reduction and the total path length traveled by the robots for
different $\alpha$ values in
Figure~\ref{fig:entropy_and_path_length}. We choose the case $r = 2$
and $\sigma^m = 2$ to highlight this point. As seen in
Figure~\ref{fig:entropy_and_path_length}a, it is evident that
increasing the $\alpha$ values reduces the number of iterations
required to reduce entropy. But, on the other hand
Figure~\ref{fig:entropy_and_path_length}b shows that increasing the
value of $\alpha$ also significantly increases the total path traveled
by the robots (see (8,8) range plots for comparison). So in
Figure~\ref{fig:darpa_subT} and Figure~\ref{fig:uc_32r} we report the violin plots for cost
defined as the normalized sum of number of iterations required for
entropy reduction and total path length.

From Figure~\ref{fig:darpa_subT} and~\ref{fig:uc_32r}, the general trend is that
heterogeneous groups of robots with some robots having $\alpha > 1$
tend to do better than other combinations. But these cases also give
rise to higher variance in the results. It also important to note that
this heterogeneity works best in smaller sensing radius and larger
sensing noise case. When sensing radius is high and sensing noise is
low, the different cases have comparable performance (compare
Figure~\ref{fig:darpa_subT}a and Figure~\ref{fig:darpa_subT}f; similarly Figure~\ref{fig:uc_32r}a and Figure~\ref{fig:uc_32r}f).  Thus
in general, it is beneficial to have robots encoded with $\alpha > 1$
with robots possessing $\alpha < 1$ to balance exploration time and
distance traveled.

\section{Conclusion and Future Work}
\label{sec:conclude}
We proposed a novel algorithm for multi-robot exploration based on a
distributed frontier assignment. We used heterogeneous BE to evaluate
frontiers and used a distributed game-theoretic framework to assign
these. We found that a team of robots with heterogeneous behaviors
outperforms a team of homogeneous robots with respect to time and
distance costs. This allowed us to heuristically prescribe BE
parameters for robot teams to efficiently explore maps.
  
In the future, we would like to understand how to dynamically change
the behaviors for more efficient assignments, and perform experiments
in challenging real-world settings.

\bibliographystyle{IEEEtran}

\appendices

\section{Proofs of results}
\label{sec:proofs}

\paragraph*{\textbf{Proof of Theorem~\ref{thm:pbrag_converge}}}

Consider a arbitrary but fixed $q \in \tsk$ and the dynamics~\eqref{eq:dist_pbrag}. 
First, notice that the arguments in~\cite[Lemma 6.1]{NM-MK-SM:25-tac} can be extended to periodically strongly connected graphs to conclude that $\forall k \in \intpos$, $\forall i \in \agt$, $M_i^q(t) = \max_{i \in \agt}z_i^q(kT)$, $S_i^q(t) = \submax_{i \in \agt}z_i^q(kT)$, $\forall t \in \{kT+2\tau,\cdots,(k+1)T - 1\} \rdef \mathcal{I}(k,\tau)$. Denote $\bar{\mathcal{I}}(T,\tau) \ldef \{kT,\cdots,kT+2\tau-1\}$. Now,
  Consider an $\epsilon > 0$. Then, by the hypothesis, for any
  $\lambda > 0$,
  $\prob{|z_i^q(t) - \rho_i^q| - \mu^q > \epsilon} < \lambda$,
  $\forall t \geq t_0$ for some $t_0$. 
Then, with probability at least $1-\lambda$, $\forall k \in \intpos$ such that $kT > t_0$, $\{w_i^q(t)\}_{t \in \mathcal{I}(k,\tau)}$ is increasing (resp. decreasing) if $i \in \argmax_{j \in \agt} \rho_i^q \rdef i^*_q$ (resp. $i \notin \argmax_{j \in \agt} \rho_i^q$). This is because of~\eqref{eq:band_converge}. 
Moreover, since $\beta_i^q(k) \to \infty$, $\forall k \in \intpos$ such that $kT > t_0$, $w_i^q(kT+T-1) = 1$ (resp. $ = 0$) if $i = i^*_q$ (resp. $i \neq i^*_q$). Finally, since $\alpha_i^q(k) \to 0$, $\min \{w_{i^*_q}^q(t)\}_{t \in \bar{\mathcal{I}}(k,\tau)} \to 1$ and $\max \{w_j^q(t)\}_{t \in \bar{\mathcal{I}}(k,\tau)} \to 0$, $\forall j \neq i^*_q$.
Thus, $\prob{\|\w_i(t) - \wb_i\| > \epsilon} < \lambda$ with $(\{q \in \tsk_i \,|\, \bar{w}^q_i = 1\})_{i \in \agt} \subseteq \opt(\agt,\tsk,\{\rho_i(q)\}_{q \in \tsk,i\in \agt})$, since $q \in \tsk$ has a unique dominating agent, completing the proof.
 Then,
 from~\cite[Theorem 6.4]{NM-MK-SM:25-tac}, it follows that
 starting from $(\w_i(t))_{i \in \agt}$, there is $ t_1 \geq t_0$
 such that $\prob{\|\w_i(t) - \wb_i\| > \epsilon} < \lambda$,
 $\forall t \geq t_1$.  
Moreover,
$(\{q \in \tsk_i \,|\, \bar{w}^q_i = 1\})_{i \in \agt} \subseteq
\opt(\agt,\tsk,\{\rho_i(q)\}_{q \in \tsk,i\in \agt})$, since each
$q \in \tsk$ has a unique dominating agent, completing the proof.

\proofend

\paragraph*{\textbf{Proof of Theorem~\ref{thm:dr_dpbrag_convergence}}}
  Suppose the hypothesis is true. From Theorem~\ref{thm:dro_bound}, 
\begin{align*}
  \Pr \Big\{ \inf_{\mathbb{Q} \in \mathcal{B}_\varepsilon(\hat{\bbp}_{N_i^t}^{i,q})}
  \hspace{-1.5ex} \mathbb{E}_{\xi \sim \bbq}[\xi] \leq \mathbb{E}[R_i^q] \leq \hspace{-1.5ex}
  \sup_{\bbq \in \mathcal{B}_\varepsilon(\hat{\bbp}_{N_i^t}^{i,q})} \hspace{-1.5ex}
  \mathbb{E}_{\xi \sim \bbq}[\xi] \Big\} \geq 1 - \theta\,,
\end{align*}
where $\mathbb{E}[R_i^q] = \mathbb{E}_{\xi \sim \bbp^*}[R_i^q]$. Moreover, using~\eqref{eq:finite_bound}, 
the samples $z_i^q(t)$, generated using~\eqref{eq:dr_sample_update}, 
converges in
probability to a $\mu^q$ band around $\mathbb{E}[R_i^q]$, with $\mu^q$
satisfying~\eqref{eq:band_converge}. 
The claim now follows from
Theorem~\ref{thm:pbrag_converge}.
\proofend

\paragraph*{\textbf{Proof of Lemma~\ref{lem:alg_conv}}}
Since entropy is a concave function,
\begin{align*}
	H_1(q) - H_1(p) \leq H_1'(p) . (q-p), \quad \forall p,q \in (0,1)\,.
\end{align*}
Further, since it is also differentiable,
\begin{align*}
	H_1'(p) = \log(1-p) - \log(p)\,.
\end{align*}
Thus, with $q = p_{k_i^*}(t+1)$ and $p = p_{k_i^*}(t)$, we get
\begin{align*}
	& H_1(p_{k_i^*}(t+1)) - H_1(p_{k_i^*}(t)) \leq \Big(\log(1-p_{k_i^*}(t))- \\
	& \hspace{9em} \log(p_{k_i^*}(t)) \Big) \Big( p_{k_i^*}(t+1) - p_{k_i^*}(t) \Big)\,.
\end{align*}

Moreover, from~\eqref{eq:het_prob_update} and~\eqref{eq:good_mapping} 
we have that
\begin{align*}
	& |p_{k_i^*}(t+1) - p_{k_i^*}^*| = |\belief(p_{k_i^*}(t)) - p_{k_i^*}^*| \\
	& \leq \mapconst\,|p_{k_i^*}(t) - p_{k_i^*}^*| \leq \cdots \leq \mapconst^t|p_{k_i^*}(0) - p_{k_i^*}^*| \leq \frac{1}{2}\mapconst^t \,.
\end{align*}
The last inequality is because $|p_k(0) - p_k^*| < 0.5$. Thus $|p_{k_i^*}(t) - p_k^*| < 0.5$, $\forall t \in \integernonnegative$. Thus,
\begin{align*}
	& -\frac{1}{2}\mapconst^t \leq p_{k_i^*}(t+1) - p_{k_i^*}^* \leq \frac{1}{2}\mapconst^t, \\
	\implies & -\frac{1}{2}\mapconst^t -(p_{k_i^*}(t) - p_{k_i^*}^*) \leq p_{k_i^*}(t+1) - p_{k_i^*}(t) \\
	& \hspace{5em} \leq \frac{1}{2}\mapconst^t -(p_{k_i^*}(t) - p_{k_i^*}^*)
\end{align*}
Now, there can be two cases. 

\emph{Case (i):} $p_{k_i^*}(t) > 0.5$. Then, $p_{k_i^*}(t+1) > p_{k_i^*}(t)$ and $\big(\log(1-p_{k_i^*}(t))- \log(p_{k_i^*}(t)) \big) < 0$. Then,
\begin{align*}
	& \Big(\log(1-p_{k_i^*}(t))- \log(p_{k_i^*}(t)) \Big) \Big( p_{k_i^*}(t+1) - p_{k_i^*}(t) \Big) \\
	& \leq \big(\log(1-p_{k_i^*}(t))- \log(p_{k_i^*}(t)) \big) \Big( -\frac{1}{2}\mapconst^t -(p_{k_i^*}(t) - p_{k_i^*}^*) \Big)
\end{align*}
\emph{Case (ii):} $p_{k_i^*}(t) < 0.5$. Then, $p_{k_i^*}(t+1) < p_{k_i^*}(t)$ and $\big(\log(1-p_{k_i^*}(t))- \log(p_{k_i^*}(t)) \big) > 0$. Then,
\begin{align*}
	& \Big(\log(1-p_{k_i^*}(t))- \log(p_{k_i^*}(t)) \Big) \Big( p_{k_i^*}(t+1) - p_{k_i^*}(t) \Big) \\
	& \leq \big(\log(1-p_{k_i^*}(t))- \log(p_{k_i^*}(t)) \big) \Big( \frac{1}{2}\mapconst^t -(p_{k_i^*}(t) - p_{k_i^*}^*) \Big)
\end{align*}
Combining these gives the result. 
\proofend

\paragraph*{\textbf{Proof of Proposition~\ref{prop:het_beh}}}
  Consider Figure~\ref{fig:entropy_band}, where we illustrate the
  allocation is distinct for two robots.  First, note from
  Figure~\ref{fig:entropy_band} that when $d_m = d_M =1$, the entropy
  values for different $\alpha_i$'s are the bold lines. Observe that,
  no matter what the $p$ values are, because of the separation in the
  bold line entropy values, there is a distinct allocation of
  cells among the robots.
  Now, in general when $d_m \leq 1 \leq d_M$, the separation appears
  after $\bar{p}$, as shown in Figure~\ref{fig:entropy_band}. Thus the
  same argument holds.  Finally note that~\eqref{eq:good_mapping}
  makes $\big([0,1] \setminus [\bar{p}, 1-\bar{p}] \big)^{|\grid|}$
  invariant under~\eqref{eq:het_update}. 
  Thus the argument can be made
  recursively. This completes the proof.
  \proofend

\end{document}